\pdfoutput=1
\documentclass[11pt]{article}

\usepackage[]{acl}

\usepackage{times}
\usepackage{latexsym}

\usepackage[T1]{fontenc}
\usepackage[utf8]{inputenc}

\usepackage{microtype}

\usepackage{multirow}
\usepackage{adjustbox}
\usepackage{subfigure}
\usepackage{pifont}
\usepackage{CJKutf8}
\usepackage{tabularx}
\usepackage{xcolor}
\usepackage{lipsum}
\usepackage[super]{nth}
\usepackage{amsmath}
\usepackage{booktabs}
\usepackage{multirow}
\usepackage[normalem]{ulem}
\useunder{\uline}{\ul}{}
\title{HUE: Pretrained Model and Dataset for \\Understanding Hanja Documents of Ancient Korea}

\author
{
    Haneul Yoo$^1$, Jiho Jin$^1$, Juhee Son$^1$, JinYeong Bak$^2$, Kyunghyun Cho$^3$, Alice Oh$^1$ \\
    $^1$KAIST, South Korea, $^2$Sungkyunkwan University, South Korea, $^3$New York University, USA \\
    \texttt{\{haneul.yoo, jinjh0123, sjh5665\}@kaist.ac.kr}, \\
    \texttt{jy.bak@skku.edu}, \texttt{kyunghyun.cho@nyu.edu}, \texttt{alice.oh@kaist.edu}
}

\begin{document}
\maketitle
\begin{abstract}
% \lipsum[1]
% \haneul{haneul} \jiho{jiho} \juhee{juhee} \alice{alice} \jy{jinyeong} \kh{kyunghyun}

% \haneul{
% Historical records in Korea before the \nth{20} centuries were primarily written in hanja, which is a dead language.
% \juhee{Even} contemporary Korean speakers cannot understand those documents at all without any support of additional information such as written ages and named entities.
% In this paper, we release the Hanja Understanding Evaluation dataset consisting of king prediction, topic classification, named entity recognition, and summarization task, which aims to extract meaningful information from hanja documents.
% % meaningful? understandable?
% We also present BERT-based models re-trained on Annals of the Joseon Dynasty and Diaries of the Royal Secretariats, corpora written in that era with hanja.
% % Our models outperform other baselines on all tasks, implying the necessity for language models pretrained on corpora fit for the time.
% % Moreover, we conduct zero-shot experiments with Daily Records of the Royal Court and Important Officials, which has never been introduced in NLP.
% Our models outperform other baselines on all tasks, even on zero-shot experiments with Daily Records of the Royal Court and Important Officials.
% To the best of our knowledge, this corpus is firstly used in our work.
% In addition, we discuss the changes over time in the Annals of the Joseon Dynasty and explore them by experiments given written ages as input.
% }

Historical records in Korea before the \nth{20} century were primarily written in Hanja, an extinct language based on Chinese characters and not understood by modern Korean or Chinese speakers. Historians with expertise in this time period have been analyzing the documents, but that process is very difficult and time-consuming, and language models would significantly speed up the process. Toward building and evaluating language models for Hanja, we release the Hanja Understanding Evaluation dataset consisting of chronological attribution, topic classification, named entity recognition, and summary retrieval tasks.
%which aims to extract meaningful information from hanja documents.
% meaningful? understandable?
We also present BERT-based models continued pretraining on the two major corpora from the \nth{14} to the \nth{19} centuries: the Annals of the Joseon Dynasty and Diaries of the Royal Secretariats.\thinspace\footnote{All codes, models, and dataset are available at \url{https://github.com/haneul-yoo/HUE.git}}
% Our models outperform other baselines on all tasks, implying the necessity for language models pretrained on corpora fit for the time.
% Moreover, we conduct zero-shot experiments with Daily Records of the Royal Court and Important Officials, which has never been introduced in NLP.
We compare the models with several baselines on all tasks and show there are significant improvements gained by training on the two corpora. Additionally, we run zero-shot experiments on the Daily Records of the Royal Court and Important Officials (DRRI).
The DRRI dataset has not been studied much by the historians, and not at all by the NLP community.
%Finally, we discuss changes over time in the Annals of the Joseon Dynasty and explore them by experiments given written ages as input.

% Classics in ancient or medieval Korean are written in hanja which is a deprecated language. Translating those documents requires enormous time and human power of experts. Providing preliminary information to non-translated ancient documents might both support experts to translate those manuscripts and act as biblographical introductions that might be leveraged by public. In this paper, we introduce Hanja benchmark and language models pretrained on hanja corpora written in the same ages. We believe that this is the first work for attributing ancient documents in hanja. Hanja benchmark is a collection of four tasks attempting to comment non-translated hanja documents and provide preliminary information: King Prediction, Topic Classiciation, Named Entity Recognition, and Summarization. We analyze how named entities and words change over time in the Annals of the Joseon Dynasty and show the correlation between written ages and our tasks emphasizing the importance of King Prediction task. Our models outperform other baselines and work well on unseen documents, implying the necessity for language models pretrained on corpora fit for the time. Our codes and Hanja benchmark are available at \url{github}.
\end{abstract}

\section{Introduction}
% (중요) 독자들이 1) 이후 섹션들을 읽는 노력을 할만큼 가치가 있도록 동기부여 시키고, 2) 이후 섹션들을 읽기 위해 준비를 시켜야함. abstract의 해설 + 우리 논문이 기여한바 정리 + 논문의 섹션 구성 미리 소개

% 

\begin{table}[!ht]
\begin{tabularx}{\columnwidth}{@{}l|l@{}}
\toprule
\textbf{Language}   & \textbf{Sentence}                                                                                           \\ \midrule
Hanja               & \begin{CJK*}{UTF8}{bsmi}\colorbox{orange}{上}\colorbox{cyan}{御}\colorbox{yellow}{經筵}。 \end{CJK*}             \\ \midrule
Modern Korean       & \begin{CJK}{UTF8}{mj}\colorbox{orange}{임금}이 \colorbox{yellow}{경연}에 \end{CJK} \\
                    & \begin{CJK}{UTF8}{mj}\colorbox{cyan}{나아갔다}. \end{CJK} \\ \midrule
Simplified Chinese  & \begin{CJK*}{UTF8}{gbsn}\colorbox{orange}{国王}\colorbox{cyan}{参加}了\colorbox{yellow}{皇家讲座}。  \end{CJK*}       \\ \midrule
Traditional Chinese & \begin{CJK*}{UTF8}{bsmi}\colorbox{orange}{國王}\colorbox{cyan}{參加}了\colorbox{yellow}{皇家講座}。 \end{CJK*}        \\ \midrule
English             & The \colorbox{orange}{King} \colorbox{cyan}{attended} \\
                    & the \colorbox{yellow}{Royal Lecture}.        \\ \bottomrule
\end{tabularx}
\caption{Example sentence in AJD}
\label{tab:hanja_example}
\end{table}

Large-scale historical records in Korea were mostly produced during the Joseon dynasty (1392-1897), 
%and preserved with uncountable contributions for centuries.
and the Institute for the Translation of Korean Classics (ITKC) keeps a comprehensive database of Korean classics at a scale of approximately 9 billion characters.
This digital archive is a great resource for Korean historians, but the documents remain in the original Hanja language \footnote{Hanja is a set of characters (script) used in ancient Korean, while Hanmun is a writing style (language) in the same era. However, we will refer to Hanja as a language following the conventions of the previous works.}. Hanja is an extinct language, and as Table \ref{tab:hanja_example} illustrates with a simple sentence, Hanja is lexically and syntactically different from modern Korean, as well as simplified and traditional Chinese.
%Precise understanding of the historical documents is required to analyze historical events and culture and convey their contents to the future generations.
%However, those documents were written in an extinct language Hanja, so present-day Korean speakers cannot understand them at all.  
%Thus, those hanja documents should be translated or at least accompanied by some meaningful information, such as written ages, topic, named entity, and \haneul{summary}. 
% \alice{
Understanding the documents in the digital archive is thus difficult and would benefit greatly from a Hanja language model which can also be used to accelerate the expert translation \cite{Gato2015anthology}.
%These additional information helps to accelerate translation process for experts. \citet{Gato2015anthology} reported that presenting some preliminary information in anthology is useful to translate literary texts professionally.
There are two corpora we can use to train the language model, the Annals of Joseon Dynasty (AJD), first introduced to the NLP community in \citep{Bak2015Mining}, and the Diaries of the Royal Secretariats (DRS) \cite{Kang2021Restoring}.
% }
%Finally, by providing some useful information related to the ancient historical corpus, the public without any background knowledge can understand the text and the translation experts increase the efficiency of whole translating process.

In this paper, we provide the HUE (\textbf{H}anja \textbf{U}nderstanding \textbf{E}valuation) dataset consisting of chronological attribution, topic classification, named entity recognition and summary retrieval, a suite of tasks to help build and evaluate the Hanja language model.
%By providing these tasks with Hanja historical data, HUE encourages training Hanja language models that help to analyze the historical documents.
In addition to AJD and DRS, we also work with the Daily Records of the Royal Court and Important Officials (DRRI).
Unlike AJD and DRS that have been analyzed by historians and contain their annotations, DRRI lacks such systematic analysis, and we use it for zero-shot learning and introduce it to the NLP community.
% Especially, the DDRI corpus used for the first time in our works.

We also provide pretrained language models (PLMs) for Hanja trained on AJD and DRS, fine-tuned for each task in HUE.
Our pretrained models on the corpora from that era outperform the existing language models built for ancient Chinese, confirming the need for specially-trained Hanja language models.
We also run additional experiments based on the analyses of entity- and word-level changes on AJD by controlling input conditions by masking named entity and giving the time period as input.
Finally, we demonstrate the effectiveness of our Hanja language model for analyzing unseen documents, running zero-shot experiments for chronological attribution and named entity recognition tasks on DRRI.
% \juhee{Zero-shot experiments for chronological attribution and named entity recognition task are conducted on DRRI dataset and the performance demonstrates that the effectiveness of Hanja language model for analyzing even unseen documents.}
% For the AJD, we provide discussion of entity/word changes over time and conduct additional experiment that gives written ages as a supplementary input to figure out the impact of written ages on model's prediction. 
% Moreover, 

Our main contributions are as follows:
\begin{itemize}
\item{
% \haneul{
% We provide the HUE dataset and Hanja PLMs to help understanding ancient Korean documents.
% }
% We provide the HUE dataset and Hanja PLMs which aim to support constructing the Hanja language models which help analyzing the historical documents in Korea.
We release the HUE dataset and Hanja PLMs to support historians to understand and analyze a large volume of historical documents written in Hanja.
% We provide the HUE Dataset and hanja PLMs to provide preliminary information for the ancient Korean documents.
To the best of our knowledge, this is the first work proposing Hanja language models and releasing a NLP benchmark dataset for ancient Hanja documents.
}
\item{
% \haneul{
%\juhee{by controlling the input condition by entity masking or given additional input such as document age we can boost up the Hanja language model performance on our task.}
% \juhee{
% We observe that some specific input conditions such as entity and document age affect to the Hanja language model performance on HUE task. 
We demonstrate that providing key information such as named entity and document age as input improves the performance of Hanja language model on the HUE tasks.
}
% \juhee{We observe that the Hanja language model performance on our task can be increased by entity masking or adding document age to the input}
% We observe some input conditions that can boost up Hanja language models by analyzing lexical features in AJD.
% }
% We diverse the input condition on various tasks and examine the performance to figure out which information in the record affects the performance of the language model.
% } 
\item{
% We run zero-shot experiments that show HUE and Hanja PLM help to analyze the unseen historical document.
We run zero-shot experiments on several HUE tasks from DRRI which have not been discussed in the NLP community, and demonstrate the performance of our Hanja language models on unseen historical documents.
}
\end{itemize}

\section{Background}
\subsection{Hanja}

% 주희 comment 
% 이 문단에서 보여줘야 할 것은 
% 한자가 고대 한국에서 쓰이던 글자이며 많은 고대 한국의 기록물들이 한자로 쓰임 
% 한자는 고대 중국어에서 유래했으나 한국어에 맞게 변화하였기 때문에 고대/현대 중국어와는 문법적, 단어적 차이가 있다
% 예시 
% 이런 언어적 차이로 인해 기존의 anchibert나 chinese bert로는  한자 기반 NLU Task에서 최적의 성능을 낼 수 없다. 
% 따라서 우리는 새로운 hanja-PLM을 공개하고자 한다 밑은 학습에 사용된 hanja corpus 에 대한 설명이다. 

\begin{CJK*}{UTF8}{bsmi}
Hanja, the writing system based on ancient Chinese characters, was the main writing system in Korea before the \nth{20} century, while Hangul, the unique Korean alphabet, has been the main writing system in Korea from the last century.
Formal records from the Joseon dynasty (1392-1897) are written in Hanja, while spoken language and some written documents were in Hangul, developed in the 15th century.
This co-existence of the written and colloquial languages has led Hanja to evolve to have the basic syntax of classical Chinese, mixing with the lexical, semantic, and syntactic characteristics of colloquial Korean.
% For example, Korean historical documents in Hanja used to transform and omit causative verb (令), and generally use the modal predicate (如何) which was rarely used in the ancient Chinese, by re-interpreting certain vocabularies or structures as their mother tongue \citep{Jeong2015Nongamjinjeok, Im2016HanjaStructure}.
\end{CJK*}

% \begin{CJK*}{UTF8}{bsmi}
% \haneul{
% Modern Korean speakers use Hangul which is an independent and unique alphabet of Korea, created by King Sejong of the Joseon Dynasty.
% Even after the creation of Hangul, most of the historical records in Korea before the \nth{20} century, especially the formal records in the Joseon dynasty, used to be written in Hanja.
% % Many society in ancient East Asia, including Japan, Vietnam, and Korea, were \emph{diglossia}, whose official language and colloquial language were different.
% Hanja, characters used in ancient Korea derived from ancient Chinese characters, became one of the linguistic variants whose basic syntactic model originated from Classical Chinese, while it had been changed and inhered the lexical, semantic, and syntactic characteristics of mother tongue of colloquial Korean.
% For example, Korean historical documents in Hanja used to transform and omit causative verb (令) and generally use the modal predicate (如何) which was rarely used in the ancient Chinese, by re-interpreting certain vocabularies or structures as their mother tongue \citep{Jeong2015Nongamjinjeok, Im2016HanjaStructure}.
% }
% \end{CJK*}

% Table \ref{tab:hanja_example} shows an example sentence frequently used in the Annals of the Joseon Dynasty.
Hanja is significantly different from both modern Korean and modern Chinese.
Modern Korean uses a different alphabet and structure, and traditional Chinese shares some characters with Hanja, while the lexicon has evolved greatly to reflect the temporal, geographical, and cultural differences between the Joseon dynasty and modern-day China.
Simplified Chinese, the current written language in China has diverged more because of the simplification of the characters.
These differences between Hanja and other related languages would lead to suboptimal performance when adopting the current Chinese language models to NLU tasks for the Korean historical Hanja documents. 

\subsection{Dataset}

\begin{table*}[!ht]
\centering
\begin{tabular}{@{}c|r|c|c|c|c@{}}
\toprule
Dataset & \multicolumn{1}{c|}{Size}  & Training data     & \multicolumn{1}{l|}{Downstream Tasks} & Zero-shot & King                                    \\ \midrule
AJD     & 230K  & \ding{52} & CA, TC, NER                          & -         & Taejo (\nth{1}) - Sunjong (\nth{27})    \\
DRS     & 1,380K & \ding{52} & -                                    & -         & Injo (\nth{16}) - Sunjong (\nth{27})    \\
DRRI    & 426K  & -         & SR                                   & CA, NER   & Yeongjo (\nth{21}) - Sunjong (\nth{27}) \\ \bottomrule
\end{tabular}
\caption{Source corpora chosen for building HUE dataset and PLMs}
\label{tab:src_corpora}
\end{table*}

We describe the three corpora of records written in Hanja during the Joseon dynasty, whose contents and additional information such as topic and named entities are provided by historians in IKTC\thinspace\footnote{\url{https://db.itkc.or.kr/}}.\textbf{}
Table \ref{tab:src_corpora} shows the list of the Hanja corpora used.
%with meta information.

\paragraph{Annals of the Joseon Dynasty (AJD)} 
also called Veritable Records of the Joseon Dynasty, is a corpus of 27 sets of chronological records, and each set covers one ruler’s reign. AJD has been translated into Korean from 1968 to 1993 and includes relevant tags such as the named entities and dates of the documents \thinspace\footnote{\url{http://esillok.history.go.kr/}}.
We use AJD for both training our Hanja language models and building the HUE dataset of NLP tasks.
% Annals of the Joseon Dynasty (AJD), also called Veritable Records of the Joseon Dynasty, include 27 different sets of chronological records, and each set covers one ruler’s reign. The last two records of Emperor Gojong and Emperor Sunjong are not treated as AJD in general, since those records did not follow the writing traditions of AJD and were inspected and produced by the Office of Governer-General during the Japanese colonial period \footnote{\url{http://esillok.history.go.kr/}}. AJD had been translated into Korean from 1968 to 1993 with preliminary information tagged, and some records are in the modernization process for more accurate and understandable outputs.
\paragraph{Diaries of the Royal Secretariat (DRS)} is a corpus of detailed records of daily events and official schedules of the court from the first King Taejo to the last (\nth{27}) Sunjong. Many of the earlier records were lost, and we use the extant records starting from the \nth{16} King Injo. DRS is known to hold the largest amount of authentic historic records and state secrets of the Joseon Dynasty\thinspace\footnote{\label{CHA}\url{http://english.cha.go.kr/}}.
We use DRS to continue pretraining the language models.
\paragraph{Daily Records of the Royal Court and Important Officials (DRRI)} is a corpus of journals written from the \nth{21} King Yeongjo to the last Emperor Sungjong and presumably initiated from the diaries of the crown prince who became the \nth{22} King Jeongjo after he was enthroned. DRRI has official daily records from both the central and the local governments, so encompasses all events in the country and reports to the king with summaries. DRRI is known to include details and events of the late Joseon Dynasty not recorded in the AJD or DRS \thinspace\textsuperscript{\ref{CHA}}, 
% Both DRS and DRRI are fully tagged with information such as named entity and written ages, but only some of them are available in Korean since the translation process is in progress.
%DRRI includes unique topics and events that do not appear at the other datasets such as AJD and DRS, which are appropriate for zero-shot experiments.
thus making it a good corpus for zero-shot experiments.
We use DRRI for the supervised summary retrieval task and the zero-shot experiments for chronological attribution and named entity recognition.

\section{HUE Dataset}
% \jiho{In this paper, we present a set of tasks aimed for providing \juhee{helpful information for the untranslated Hanja documents}
% %meta information of the untranslated hanja documents.
% Also, we release the benchmark HUE (\textbf{H}anja \textbf{U}nderstanding \textbf{E}valuation) for evaluating the hanja modeling performance on the presented tasks.}

% The HUE (\textbf{H}anja \textbf{U}nderstanding \textbf{E}valuation) dataset, which centers on Hanja document understanding tasks, covers the broad Korean historical documents. 
% The objective of the HUE dataset is to induce development of the generalizable Hanja language understanding system so that the model can perform well on arbitrary historical documents written in Hanja.
% % The information produced by HUE such as 
% \juhee{ The models trained with the NLU tasks in HUE can predict and generate meaningful information such as the age of the document that helps to research historical records. }
    % 모델이 historical document의 분석/이해 도움이 되는 정보들을 학습할 수 있도록  구성되었다..
    % This dataset is configured so that the model can learn historical document analysis/understanding helpful information.
    % the model can figure out the important information of the historical data.
% Therefore, we expect that the language models based on the HUE will ultimately help the historians to analyze ancient historical documents. 
    % tasks 들은 실제 전문가들이 고대 문서를 해석하는데 도움이 되기도 함. 
    % model이 임의의 hanja document에서도 좋은 성능을 내도록 task 를 구성하였다. 
    % we establish the tasks such that good performance 
    % 

The HUE (\textbf{H}anja \textbf{U}nderstanding \textbf{E}valuation) dataset is built to assist history scholars to understand Korean historical records written in Hanja.
%It covers comprehensive historical documents so that language models evaluated on this dataset can perform well on arbitrary Hanja documents.
%HUE suggests tasks providing understandable and explainable information related to the documents.
HUE consists of chronological attribution, topic classification, named entity recognition, and summary retrieval, which are tasks that can provide helpful information for studying the documents.
We expect that the language models based on HUE will ultimately help historians to interpret unseen historical documents and public to grasp basic concept of those documents.
% \subsection{Design Principles}
% We design this benchmark with the following principles:
% \begin{itemize}
%     \item \textit{Improving efficiency in translation:} To provide \juhee{additional} information that can be efficiently used in the translation process of experts
%     \item \textit{Helpful to understand hanja documents:} To provide \juhee{useful} information delivered in bibliographical introductions which are documents provided by translators or analysts to help public to understand the document.
% \end{itemize}
%We construct the HUE dataset with predicting document age, topic, named entity, and summary for the Hanja documents. 
%Topic, named entity, and summary will gives blueprints of Hanja documents to the experts for translation.
%Document age prediction will gives clue to find relevant documents and refer to historical facts, while most documents lack of that information.
% \juhee{Document age prediction is essential for ancient documents because it helps to find relevant documents or refer to the historical facts, but most of the documents lack of the written period information.}
We describe each task in detail below.

\subsection{Task Description}
\label{sec:task}
\paragraph{Chronological Attribution (CA)}
is a classification task predicting the ruling king when the document was written.
% \juhee{the king of the era}
When given a Hanja document from AJD, a classifier outputs one of the 27 kings of the Joseon dynasty.
% chronological attribution (CA) is a single sentence classification task to predict which king reign the hanja document was written.
% In CA, given a hanja document among the articles of AJD, a text classifier should predict a king among 27 kings in the Joseon dynasty. 
Chronological attribution of the undiscovered document is the first step in anthology to interpret and translate it.
Korean historians mostly divide the history of the Joseon Dynasty based on the reigning king, so that we treat chronological attribution as a classification task.

\paragraph{Topic Classification (TC)}
is a multi-class and multi-label classification task to find the topics of the given document.
For TC, we use Hanja document from AJD.
% The original data have 3-level topics, but we exclude the most detailed level with 199 classes. Instead, 
We suggest two levels of topics, namely major and minor categories.
The major categories consist of 4 broad topics: politics, economy, society, and culture.
The minor categories go with 106 sub-topics such as diplomacy, agriculture, and science.
% Topic Classification (TC) is a multi-class and multi-label classification task with a single sentence classification task to find appropriate topics or subjects.
% TC consists of three divisions with each different number of classes.
% The major categories, the broadest task, provides four labels: political, economy, social, and culture, while the minor categories goes with 106 classes such as diplomacy, agriculture, and science describing sub-topics of the major categories.
% The input source of TC is hanja articles of AJD as well as CA.
% The original data include 3-level topics, but we exclude the most detailed level which contains 199 classes due to some classes with only few number of samples.

\paragraph{Named Entity Recognition (NER)}
is a sequence tagging task, 
%
% Named Entity Recognition (NER) is a sequence tagging task to find the position of the named entity used in the input text and identify the entity type.
identifying the two types of named entities, person and location, from the Hanja document from AJD.
We divide train, validation, and test sets such that there are no overlapping entities across the sets.

\paragraph{Summary Retrieval (SR)}
is a task to find the most relevant summary that matches the content among the summary candidates.
For this task, we use DRRI, in which each document is a pair of summary (\emph{gang}) and detailed content (\emph{mok}).
% Among 426K DRRI articles, 265K articles are included which have both \emph{gang} and \emph{mok}. 
Among 426k articles, 265k articles in DRRI dataset contain both \emph{gamg} and \emph{mok}.
Also, we exclude those with \emph{gang} longer than \emph{mok}, in which \emph{gang} is not the summary of \emph{mok}.
%Finally, we have 213K pairs of the summary and the content.
The final dataset contains 213K pairs of content and the corresponding summaries.
We describe more details of the preprocessing in the Appendix.

% Summarization is a retrieval task to find the most relevant summary that matches the content among the summary candidates.
% In summarization, we use DRRI as a summarization dataset, which is written in a Gangmok-style, instead of AJD.
% DRRI consists of gang (summary) and mok (detailed contents).
% DRRI consists of some articles whose gang is not precisely a summary of mok, especially at the infancy phase in the reign of King Yeongjo, since DRRI inherits from the diaries of the crown prince.
% We adopt the naive way of comparing the length between gang and mok to exclude those examples and get 216K articles for summarization.

% We fine-tuned PLMs as a re-ranker to solve the summarization task as a retrieval problem. First, we retrieve top-100 relevant gangs (summaries) with BM25 \cite{Robertson2009BM25}, and then we fine-tuned BERT-based models as a binary classifier to determine whether the summary matches the content. We fine-tuned those models with top-12 negative examples retrieved by BM25. Finally, we re-rank the summaries with respect to cross-entropy loss \cite{Nogueira2019Reranker}. If the ground truth summary is not included in the top-100 relevant summary candidates retrieved in the first step since the top-100 accuracy of BM25 was merely 52.47\%, we replace the last \nth{100} summary with the ground truth.

% 426K examples → 265K examples → 213K examples

\section{Hanja Pretrained Model}
%%% No PLM for Hanja & Related BERT models
%% AnchiBERT
%% mBERT

As far as we know, there have been no pretrained language models for the Hanja language.
One can use related LMs, the pretrained models for ancient Chinese as well as
% We believe that there is no pretrained language model for the Hanja domain. Hence, we decided to use and re-train pretrained language models for traditional Chinese or ancient Chinese. In this paper, we use two pretrained language models: AnchiBERT \cite{Tian2021AnchiBERT} and multilingual BERT \cite{Devlin2019BERT}.
multilingual BERT \cite{Devlin2019BERT} which includes traditional Chinese in its training corpus.
AnchiBERT \cite{Tian2021AnchiBERT} is pretrained in ancient Chinese with the Chinese anthologies written around 1000BC to 200BC.
% Also, we use a multilingual BERT\cite{Devlin2019BERT}, which is trained in traditional Chinese corpora in modern texts.
% Even though there is no time overlap between Hanja documents in the Joseon dynasty and the ages when train corpora of those two models are written, we assume that there will be a comparable amount of vocab overlap between the Hanja documents and the models.
There is some vocabulary overlap between the Hanja documents and traditional Chinese corpora, we can adopt multilingual BERT and AnchiBERT to learn the representations of the Hanja texts.

%%% Our Approach
%% Pretrain
%% 1) without pretraining
%% 2) AnchiBERT/mBERT
%% 3) AnchiBERT/mBERT + continuing pretraining on AJD/DRS
We propose the pretrained language models suitable for Hanja documents by continuing pretraining those two models on both AJD and DRS.
%% Ratio of Unknown Tokens
%Table \ref{tab:unk_token} shows the ratio of unknown tokens in the training corpus by each PLM.
Table \ref{tab:unk_token} shows the ratio of unknown tokens in the test set of AJD by each model.
It implies that existing AnchiBERT and multilingual BERT can also be exploited as language models for Hanja documents written in the Joseon dynasty, but the second phase of pretraining on the corpora of that era remarkably decreases the ratio of unknown tokens.
% We fine-tune all models to benchmark tasks and analyze the results.

\begin{table}[!ht]
\centering
\resizebox{\columnwidth}{!}{
\begin{tabular}{@{}l|r|r@{}}
\toprule
         & AnchiBERT                & mBERT                 \\
         & \cite{Tian2021AnchiBERT} & \cite{Devlin2019BERT} \\ \midrule
original & 0.88\%                   & 0.76\%                \\
+AJD/DRS & 0.04\%                   & 0.04\%                \\ \bottomrule
\end{tabular}
}
\caption{
Unknown token ratio of each model in test set for CA, TC, and NER task in HUE.
The first row indicates the results of the original PLMs without any additional pretraining, and the second row indicates those with continued pretraining on AJD and DRS.
}
\label{tab:unk_token}
\end{table}

% %% Finetune
% \jiho{We can apply the models to each task described in Section \ref{sec:task} by fine-tuning them.
% %% Summarization
% \haneul{
% Especially for the summary retrieval task, we fine-tune each model to act as a re-ranker in retrieval task.
% }
% % Especially for the summarization task, we fine-tune the PLMs as a re-ranker to solve the task as a retrieval problem.
% We first retrieve top-\haneul{$k$} relevant \emph{gang}s (summaries) with BM25 \cite{Robertson2009BM25} among all \emph{gang}s in the dataset.
% Then, we fine-tune the model as a binary classifier to determine whether the summary matches the content of the \emph{mok} with the cross-entropy loss \cite{Nogueira2019Reranker}.
% If the ground truth summary is not included in the top-\haneul{$k$} relevant summary candidates, we replace the last \haneul{$k^{\text{th}}$} summary with the ground truth.
% We use \haneul{$k=12$} for training and \haneul{$k=100$} for inference.
% }

%We first retrieve top-100 relevant \emph{gang}s(summaries) with BM25\cite{Robertson2009BM25}, and then fine-tune the PLMs as a binary classifier to determine whether the summary matches the content of the \emph{mok}.
%We fine-tuned those models with top-12 negative examples retrieved by BM25.
%Finally, we re-rank the summaries with respect to cross-entropy loss\cite{Nogueira2019Reranker}.
% If the ground truth summary is not included in the top-100 relevant summary candidates retrieved in the first step since the top-100 accuracy of BM25 was merely 52.47\%, we replace the last \nth{100} summary with the ground truth.

\section{Experiment}

\subsection{Experimental Settings}

%  데이터는 어떤 데이터를 쓸 것이고 baseline은 무엇이며 내가 제시하는 모델은 어떤 것이다라고요

We conduct experiments on HUE with our pretrained model.
For the baseline model, we use BERT without pretraining and compare it to various BERT models described in Section \ref{sec:task}.
%% Finetune
Specifically, we fine-tune each model to act as a re-ranker in the retrieval task for the summary retrieval.
% We can apply the models to each task described in Section \ref{sec:task} by fine-tuning them.
%% Summarization
% Especially for the summarization task, we fine-tune the PLMs as a re-ranker to solve the task as a retrieval problem.
We first retrieve top-$k$ relevant \emph{gang}s (summaries) with BM25 \cite{Robertson2009BM25} among all \emph{gang}s in the dataset.
Then, we fine-tune the model as a binary classifier to determine whether the summary matches the content of the \emph{mok} with the cross-entropy loss \cite{Nogueira2019Reranker}.
If the ground truth summary is not included in the top-$k$ relevant summary candidates, we replace the last $k^{\text{th}}$ summary with the ground truth.
We use $k=12$ for training and $k=100$ for inference.
% We use BERT without any pretraining but only with fine-tuning for each task as the baseline, and compare it to various BERT models to inspect how critical pretraining on Hanja (or Hanja-related) domain data is for the \haneul{historical} representation learning.
As the representative metrics, we present F1 scores for CA, TC, and NER tasks, and Mean Reciprocal Rank (MRR) for SR.
The detailed results with other metrics are also available in Appendix.

\subsection{Overall Results}

\begin{table*}[!ht]
\centering
\begin{tabular}{@{}l|c|cc|cc|c@{}}
\toprule
                                   & \textbf{CA}       & \multicolumn{2}{c|}{\textbf{TC}}      & \multicolumn{2}{c|}{\textbf{NER}}     & \textbf{SR}       \\
                                   &                   & \textbf{Major}    & \textbf{Minor}    & \textbf{Person}   & \textbf{Location} &                   \\ \cmidrule(l){2-7} 
                                   & F1                & F1                & F1                & F1                & F1                & mrr               \\ \midrule
BERT not pretrained                & 54.26             & 68.91             & 61.52             & 92.13             & 87.10             & 52.85             \\
mBERT \cite{Devlin2019BERT}        & 75.29             & 79.59             & 76.46             & 91.63             & 86.02             & 67.06             \\
AnchiBERT \cite{Tian2021AnchiBERT} & 75.74             & 85.81             & 75.22             & \textbf{93.28} & \textbf{88.01}    & 67.92             \\
mBERT + AJD/DRS                    & \underline{77.77} & \underline{87.13} & \underline{77.84} & 92.83             & 87.90             & \underline{73.88} \\ 
AnchiBERT + AJD/DRS                & \textbf{79.33}    & \textbf{88.33}    & \textbf{78.10}    & \underline{93.13}    & \underline{87.91} & \textbf{74.29}    \\ \bottomrule
\end{tabular}
\caption{Evaluation results of PLMs on HUE dataset}
\label{tab:overall}
\end{table*}

% We also set BERT without any pretraining as a baseline model to inspect whether pretraining language models on Hanja domain data is critical in predicting preliminary attributes.

% We use BERT without any pretraining but only with fine-tuning for each task as the baseline, and compare it to various BERT models to inspect how critical pretraining on Hanja (or Hanja-related) domain data is for the \haneul{historical} representation learning.
% As the representative metrics, we present F1 scores for CA, TC, and NER tasks, and Mean Reciprocal Rank (MRR) for SR.
% The detailed results with other metrics are also available at the Appendix.
% }
% We present all results in Table \ref{tab:overall}, using F1 score for the representative metrics except for SR using Mean Reciprocal Rank(MRR).

% It delivers a strong observation that BERT without any pretraining shows the poorest results among all tasks. Also, AnchiBERT and mBERT, which are existing PLMs on other domain, show a better result than BERT without any pretraining, while the models re-trained on Hanja documents achieve the best performance among all tasks. It indicates that all these subtasks on understanding historical documents require time-specific and domain-specific data.
Table \ref{tab:overall} shows the overall experimental results.
The bold and the underlined texts in the table specify the best and the second best result, respectively.
BERT without any pretraining shows the poorest results across all the tasks.
AnchiBERT and mBERT, which are existing language models on the relevant domains, show better results, and the models continued pretraining on Hanja documents achieve the best performance among all tasks.
This tendency indicates that all these tasks on understanding historical documents require pretraining language models on time-specific and domain-specific data.
% The detailed experimental results with other metric for all tasks are available at Appendix.

AnchiBERT pretrained on the Hanja corpora shows slightly better performance than mBERT pretrained on the same corpora.
We assume this is because the original training corpora of AnchiBERT are much closer to the Hanja documents, even the era of those two corpora are completely different.
The writing style of both Hanja documents in the Joseon dynasty and anthologies in ancient China had come from Classical Chinese and share similarities.
On the other hand, the training corpora of mBERT is a contemporary texts whose characters contains Traditional Chinese, but the structure and the format might be considerably changed.

\paragraph{Chronological Attribution (CA)}

Our models continued pretraining on Hanja corpora outperform other baselines on CA.
Detailed analysis on CA result is illustrated in Section \ref{sec:CA_analysis}.

\paragraph{Topic Classification (TC)}

\begin{figure}[!ht]
    \centering
    \subfigure{\includegraphics[width=\columnwidth]{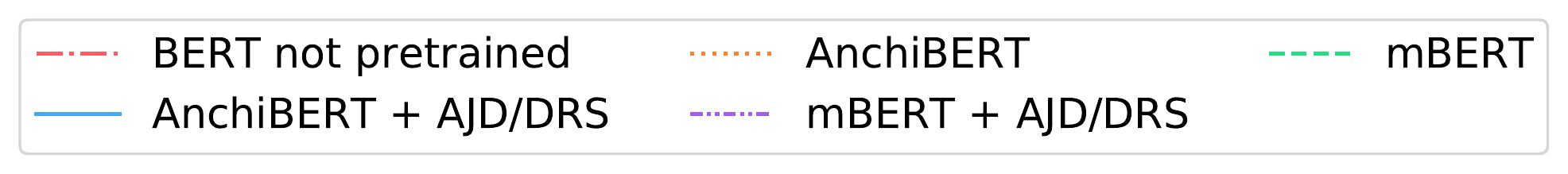}}
    \addtocounter{subfigure}{-1}
    \subfigure[\centering Div 1 (4 classes)]{\includegraphics[width=0.45\columnwidth]{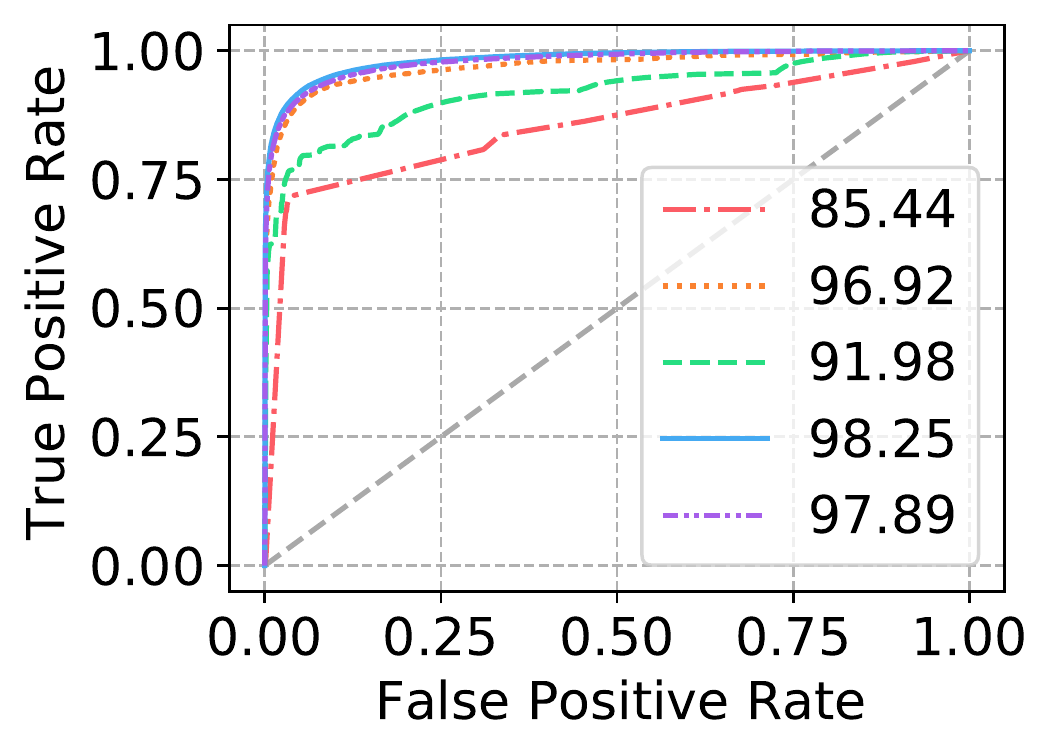}}
    \subfigure[\centering Div 2 (106 classes)]{\includegraphics[width=0.45\columnwidth]{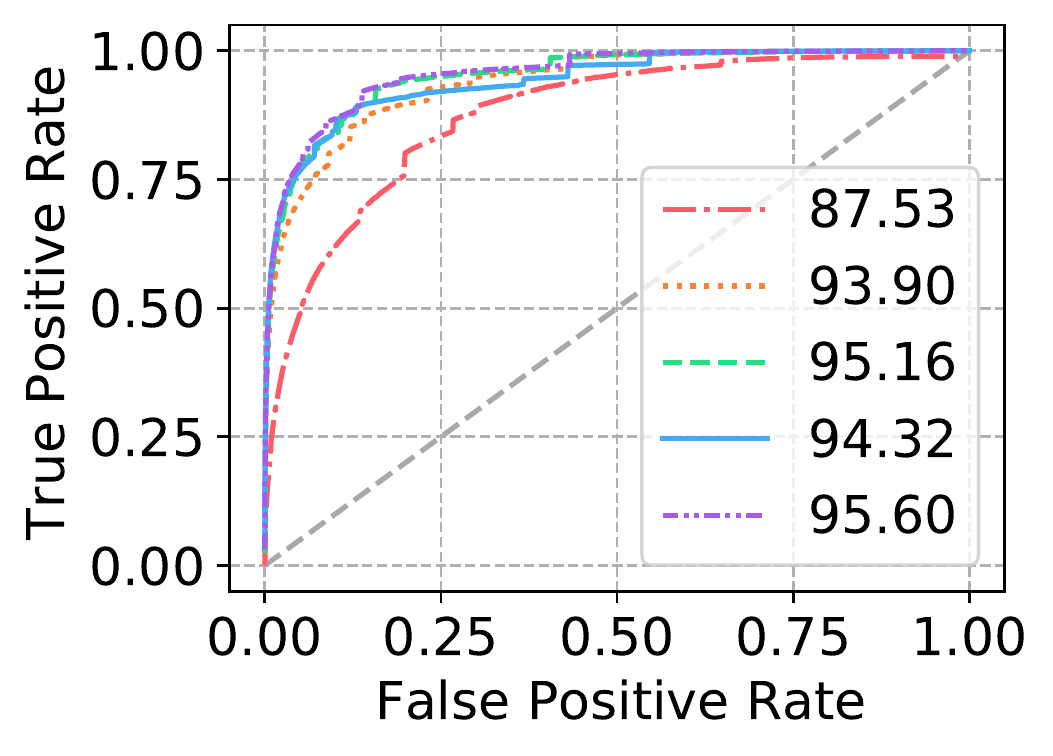}}
    % \subfigure[\centering Div 3 (199 classes)]{\includegraphics[width=0.3\linewidth]{final_figures/topic-div3.pdf}}
    \caption{ROC Curve and AUC for Topic Classification. Each value in the legend indicates the AUC score.}
    \label{fig:roc_auc}
\end{figure}

Figure \ref{fig:roc_auc} gives ROC curves and AUC values of each model on each task.
Our models show the similar trends to the overall results, outperforming other language models.
For the evaluation results including F1 score, we find and set the best threshold to each label by Youden's index \citep{Youden1950index}.

While F1 score goes down as the number of classes increases from 4 to 106, there is no significant difference on AUC value. 
This might result from consistently high recall achieving around 90\% on both tasks.
It indicates that the threshold is too low and models tend to predict plausible topics as many as they can, which might be solved by controlling the threshold.
AnchiBERT pretrained on AJD and DRS, which shows the best performance, predicts 6.39 labels in average, while the average number of ground truth labels of the minor categories is 1.97.
This is probably due to the meaning overlaps in minor categories.
For instance, minor categories such as revenue, finance, general price level, and commerce are the sub-categories of economy in the major categories, whose use case cannot be strictly distinguished.
It would be more appropriate in this case to provide all plausible topics roughly rather than suggesting the one only with high certainty.
% Since TC is a multi-label classification task whose example might have multiple labels as the answer, we measure Hamming score along with accuracy. In this case, accuracy is the exact match score, and the hamming score is the accuracy of subset matched, $\mid T \cap P \mid / \mid T \cup P \mid$, where $T$ is set of true labels and $P$ is set of predicted labels \cite{Godbole2004hammingscore}. For the evaluation results in Table \ref{tab:topic}, we find and set the best threshold to each label by Youden's index. All pretrained models outperform BERT without pretraining, and two LMs re-trained on Hanja documents show the best performances.

% Table \ref{tab:topic} shows results with all metrics except for recall drops if it goes to detailed division, while recall in all divisions achieves around 90\%. It indicates that the threshold is too low and models tends to predict plausible topics as many as they can, which might be solved by controlling the threshold. This is probably due to the meaning overlaps in minor categories. For instance, there are minor categories such as revenue, finance, general price level, and commerce are the sub-categories of economy in the major categories, whose use case cannot be strictly distinguished. It would be more appropriate in this case to provide all plausible topics roughly rather than suggesting the one only with high certainty.

\paragraph{Named Entity Recognition (NER)}

NER also indicates similar trends to the overall benchmark tasks, but with a small gap among models including BERT without pretraining.
It implies that NER in Hanja documents is a comparably easy task.
This might result from certain patterns in named entities in Hanja.
Most of the person entities are 3 letters starting with the common characters (family name), and most of the location entities end with the common characters meaning locations or buildings.
All models tend to predict person entities better than location entities.

\paragraph{Summary Retrieval (SR)}

% \begin{table}[!]
% \resizebox{\columnwidth}{!}{
% \begin{tabular}{l|ccc}
% \hline
% Model               & mrr               & Top-1             & Top-10            \\ \hline
% BM25                & 29.87             & 25.58             & 33.98             \\
% BERT not pretrained & 52.85             & 99.20             & 99.64             \\
% AnchiBERT           & 67.92             & 99.20             & \underline{99.85} \\
% mBERT               & 67.06             & 99.32             & 99.50             \\
% AnchiBERT + AJD/DRS & \textbf{74.29}    & \textbf{99.64}    & \textbf{99.91}    \\
% mBERT + AJD/DRS     & \underline{73.88} & \underline{99.44} & 99.59             \\ \hline
% \end{tabular}
% }
% \caption{Evaluation Results for our PLMs on Summairzation}
% \label{tab:summarization}
% \end{table}

All fine-tuned BERT-based re-rankers outperform BM25 whose MRR is merely 29.87\%, mostly retrieving the ground truth answer at the first trial.
Likewise, our models shows the best results, while BERT without pretraining show the lowest MRR.
It additionally implies that BERT-based re-rankers might be exploited for retrieving relevant documents from different chronicles in terms of written style or contents.

% \subsection{Analysis of Language Model Performance According to Input Condition}
% Analysis of factors influencing the performance of the language model.
% 언어모델의 성능에 영향을 주는 input 요소 분석 
% lm 성능에 영향을 주는 특정 정보 분석 
% \subsection{Boost Up Language Models}
\subsection{Effect of Entity and Document Age on Language Model}

% We conduct additional analysis on the performance change of the language model according to some input condition.
We investigate whether providing additional information as input can improve the performances of language models.
% We analyze some input conditions that affect to the language model performance.
For CA, we mask all named entities in the input and fine-tuned the language model on masked data to examine the impact of entity information. 
For TC and NER, we concatenate document age information to the input text and run comparative experiments to verify the importance of time period on historical texts.

\paragraph{Entity-Masked Chronological Attribution}

\begin{table}[!ht]
\centering
\begin{tabular}{@{}l|c@{}}
\toprule
                                  & \textbf{CA (Masked)}               \\ \midrule
BERT not pretrained                & 75.07 ($\Delta$ 20.81)               \\
mBERT \cite{Devlin2019BERT}        & 83.44 ($\Delta$ 8.15)                \\
AnchiBERT \cite{Tian2021AnchiBERT} & 82.45 ($\Delta$ 6.71)                \\
mBERT + AJD/DRS                    & \underline{83.57} ($\Delta$ 5.80)    \\
AnchiBERT + AJD/DRS                & \textbf{83.58} ($\Delta$ 4.25)       \\
 \bottomrule
\end{tabular}
% }
\caption{F1 scores on Chronological Attribution given named entities masked. The value inside parenthesis indicates the increase in performance after masking the named entities.}
\label{tab:CA_masked}
\end{table}

% \begin{table}[!]
% \centering
% \begin{tabular}{@{}l|l@{}}
% \toprule
%                                   & \textbf{CA (Masked)}              \\ \midrule
% BERT not pretrained                & 75.07 ($\Delta$ 20.81)            \\
% mBERT \cite{Devlin2019BERT}        & 83.44 ($\Delta$ 8.15)             \\
% AnchiBERT \cite{Tian2021AnchiBERT} & 82.10 ($\Delta$ 6.37)             \\
% mBERT + AJD/DRS                    & \textbf{84.06} ($\Delta$ 6.28)    \\
% AnchiBERT + AJD/DRS                & \underline{83.58} ($\Delta$ 4.25) \\ \bottomrule
% \end{tabular}
% \caption{F1 score on chronological attribution given named entities masked. The value inside parenthesis indicates the increased performance after masking the named entities.}
% \label{tab:CA_masked}
% \end{table}

Table \ref{tab:CA_masked} shows the difference on experimental results in CA when the named entities in the given input texts are masked.
Compared to the default settings which does not mask named entities, all models show significant improvements.
This is probably because models can truly focus on the content and changes in writing style without any disturbance of location entities consistently used for the whole era.
Fine-tuned models with entity masked inputs also achieve nearly the same level of performance in the inference with plain inputs including named entities.
It suggest to fine-tune models masking named entity in CA, considering the real scenario whose inference texts lack time period information.

\paragraph{Topic Classifciation and Named Entity Recognition with the Age of Document}

\begin{table*}[!]
\centering
% \resizebox{\linewidth}{!}{
\begin{tabular}{@{}l|cc|cc@{}}
\toprule
                                  & \multicolumn{2}{c|}{\textbf{TC}}                                       & \multicolumn{2}{c}{\textbf{NER}}                                      \\
                                  & \textbf{Major}                     & \textbf{Minor}                    & \textbf{Person}                   & \textbf{Location}                 \\ \midrule
BERT not pretrained                & 86.99 ($\Delta$ 18.08)               & 70.58 ($\Delta$ 9.29)               & 91.90 ($\Delta$ -0.23)              & 86.43 ($\Delta$ -0.68)              \\
mBERT \cite{Devlin2019BERT}        & \textbf{93.57 ($\Delta$ 13.98)}      & 73.64 ($\Delta$ 3.63)               & 93.71 ($\Delta$ 2.08)               & 87.84 ($\Delta$ 1.83)               \\
AnchiBERT \cite{Tian2021AnchiBERT} & 89.32 ($\Delta$ 3.51)                & 73.57 ($\Delta$ 4.27)               & \underline{94.82} ($\Delta$ 1.54)   & \underline{89.84} ($\Delta$ 1.83)   \\
mBERT + AJD/DRS                    & 90.02 ($\Delta$ 2.89)                & \textbf{74.84 ($\Delta$ 3.56)}      & \textbf{94.88 ($\Delta$ 2.05)}      & \textbf{89.88 ($\Delta$ 1.98)}      \\
AnchiBERT + AJD/DRS                & \underline{90.02} ($\Delta$ 1.69)    & \underline{74.54} ($\Delta$ 2.47)   & 94.70 ($\Delta$ 1.57)               & 89.46 ($\Delta$ 1.54)               \\
 \bottomrule
\end{tabular}
% }
\caption{F1 scores on topic classification and named entity recognition given document age. The value inside parenthesis indicates the difference on performance after providing document age.}
\label{tab:king_as_input}
\end{table*}

It is for granted to regard that historical texts written over several eras reveal the time changes with respect to lexical choices and contents.
Section \ref{sec:changes_over_time} confirms the hypothesis above in terms of $n$-grams changes over time.
% As we observe that texts in AJD implies the time changes of the Joseon dynasty, we experiment whether those changes affect other understanding tasks in fact. 
Table \ref{tab:king_as_input} shows gaps between experimental results of TC and NER given which king reigned when the document was written. 
Providing document age definitely increases the performance of classifying topics and tagging named entities. 
There was a big gap on difference with non-pretrained BERT in TC, which is probably due to the poor performance of itself in the original setting. 
All models show similar trends on both tasks with improved performances compared to the original settings without document age as input. 
It is an obvious result considering that the first step for ancient manuscript is assuming the written era.
It implies the significance of chronological attribution task in HUE, conveying that chronological attribution task might improve the performance of other HUE tasks.

\subsection{Zero-shot Experiment}

\begin{table*}[!ht]
\centering
\begin{tabular}{@{}l|c|cc@{}}
\toprule
                                   & \textbf{CA}       & \multicolumn{2}{c}{\textbf{NER}}      \\
                                   &                   & \textbf{Person}   & \textbf{Location} \\ \midrule
BERT not pretrained                & 26.94             & 68.50             & 39.86             \\
mBERT \cite{Devlin2019BERT}        & 32.19             & 72.08             & \textbf{83.36}    \\
AnchiBERT \cite{Tian2021AnchiBERT} & 30.65             & 71.85             & 77.23             \\
mBERT + AJD/DRS                    & \underline{35.28} & \textbf{88.48}    & 72.08             \\
AnchiBERT + AJD/DRS                & \textbf{35.85}    & \underline{82.86} & \underline{76.53} \\
 \bottomrule
\end{tabular}
\caption{F1 scores of chronological attribution and named entity recognition task on DRRI in zero-shot settings}
\label{tab:zero-shot}
\end{table*}

% The HUE designed for building Hanja language models that can provide some meaningful information to historians. 
Countless number of Hanja documents still remain without any analysis and new documents continue to be unearthed.
Therefore, we run zero-shot experiments to verify the effectiveness of our language models on extracting information from the historical documents irrelevant to the training corpora.
We use DRRI dataset which is not included in both pretraining and fine-tuning data of our Hanja language models and execute CA and NER.

Table \ref{tab:zero-shot} shows experimental results with CA and NER on DRRI. 
All models perform comparably well on the both tasks, regarding that random model will achieve approximately 3.70\% performances with 27 classes in CA. % 더 나은 문장으로 바꿀방안 찾기 
Also, all models in CA commonly show high precision which might be due to the monotonous and redundant phrases in the veritable records. It shows similar trends compared to the Table \ref{tab:overall}, but the gap among models was notably emphasized in the zero-shot settings. This results imply that our CA models might be exploited for the time period prediction of unseen documents in anthology with a reliable level.

Our models outperform others on NER achieving absolutely high performances, though entity maps between AJD and DRRI do not match strictly. Interestingly, all models tend to predict location entities better than person entities which is the opposite result compared to the original NER on AJD. It is probably due to the characteristics of each entity, where location entities are all commonly used in nationwide while person entity might differ by situation.
Further analysis on person and location entities in the view of time changes is described in Section \ref{sec:changes_over_time}.
We present that our models trained on the corpora of the Joseon dynasty provide reliable results on unseen records, implying that our model can be exploited for the low-resourced documents.

\section{Further Analysis}
% In this section, we provide further analysis on how time changes affect written texts in a historical view and how those changes in time affect other HUE tasks.

% \subsection{Analysis on King Prediction}
\subsection{Do Historical Events Affect Language Models?}
\label{sec:CA_analysis}

% There is no doubt that veritable records depicting daily events for several centuries leave a mark to texts. 
% In this section, we observe whether language models capture those historical cues revealed in the AJD.
% All models show similar findings, and we show the results with AnchiBERT pretrained on AJD and DRS which achieves the best performance.

\begin{figure}[!t]
    \centering
    \subfigure[Log-scale Confusion Matrix]{\includegraphics[width=0.8\columnwidth]{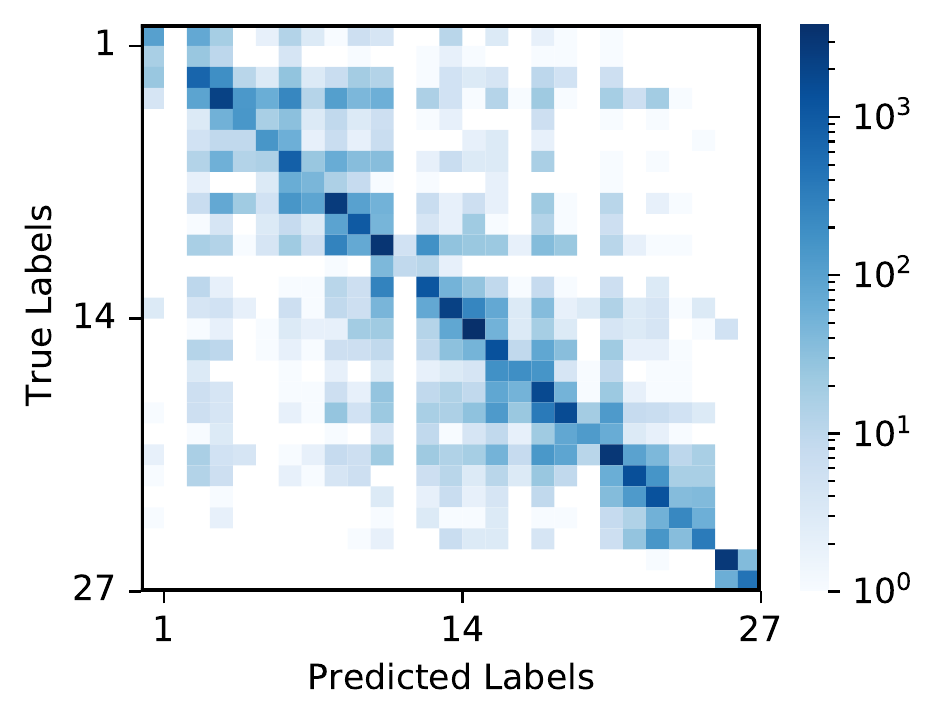}}
    \subfigure[Mean Absolute Error per King]{\includegraphics[width=0.8\columnwidth]{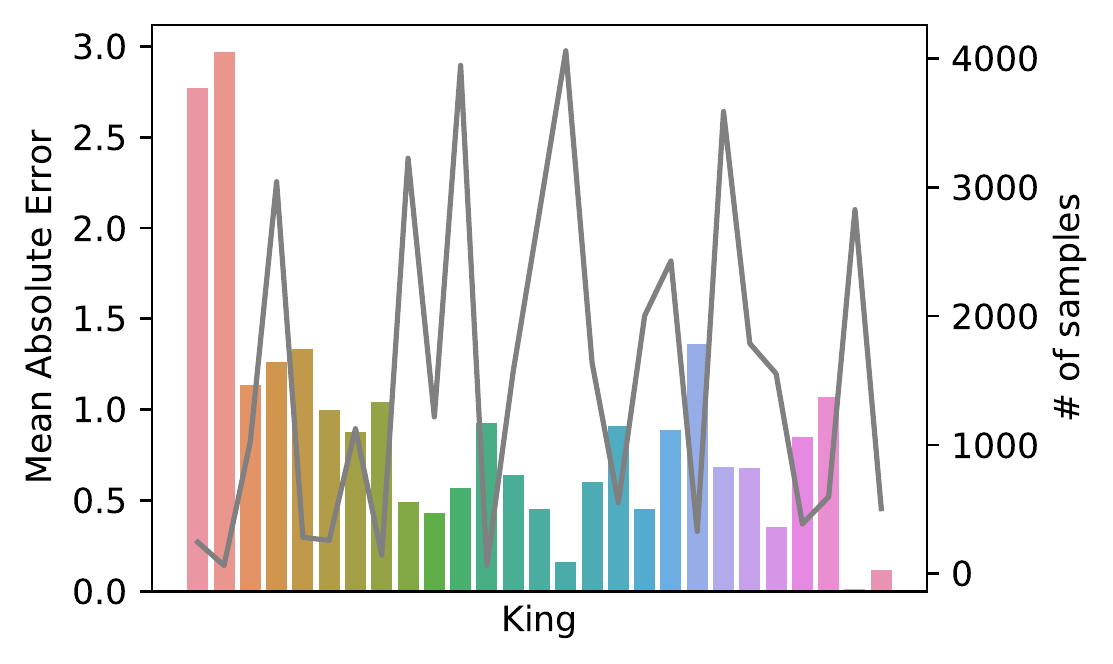}}
    \caption{Chronological attribution results with AnchiBERT + AJD/DRS. Figure (a) shows the confusion matrix. In Figure (b), each bar indicates the mean absolute error and the line indicates the number of samples for each king.}
    \label{fig:conf_diff}
\end{figure}

To figure out the effect of historical events on models' prediction, we analyzed the output of language models on CA. 
Figure \ref{fig:conf_diff} (a) shows a log-scale confusion matrix of AnchiBERT continued pretraining on AJD and DRS, and
Figure \ref{fig:conf_diff} (b) indicates the mean absolute error between the predicted king order and the ground truth per each King. 
The x-axis in the Figure \ref{fig:conf_diff} (b) means the changes of time by the king reign period.
Each bar in Figure \ref{fig:conf_diff} (b) indicates the mean absolute error between the order of ground truth king and the one of predicted, and the line graph means the number of samples in the test set.

The results of the last two Emperors, Gojong and Sunjong, are remarkable in that the model rarely gets confused with those two labels to others and tends not to fail, showing notable difference on Figure \ref{fig:conf_diff} (a) and significantly low mean absolute error on Figure \ref{fig:conf_diff} (b).
We believe that this is because our model learns the difference between those two records and the others in the historical view and get cues to distinguish them.
The last two records are not treated as AJD in general, since those records 
% did not follow the writing traditions of AJD and 
were inspected and produced by the Office of Governer-General during the Japanese colonial period with the view of Empire of Japan who ruled Korea\thinspace\footnote{\url{http://esillok.history.go.kr/}}.

The mean absolute error of the predicted order of each king achieved the difference around one, except for the first King Taejo and the second King Jeongjong, whose errors are almost doubled.
We hypothesize that this is mainly because there are too small number of examples in those classes. A similar tendency where the more samples are, the less mean absolute error be has been observed in other classes.
Also, the writing style of AJD had settled down from the third King Taejong, and those noisy records might confuse models to predict the exact dates.

\subsection{Do Time Changes Affect Written Texts?}
\label{sec:changes_over_time}

% \subsubsection{Do Time Changes affect Written Texts?}

It stands to reason that AJD written in five centuries reveal the features of the language changes.
In this section, we investigate the hypothesis above with respect to the named entities and $n$-grams.
% In this section, we investigate whether changes over time affect to the written texts in the Joseon dynasty which lasted in five centuries.
% More specifically, we ask whether named entities and words have changed as time goes by.

\paragraph{Words Change over Time}

% \begin{figure*}
%     \centering
%     \includegraphics[width=\linewidth]{figures/trigram_per_king.png}
%     \caption{Number of Overlapped trigrams per King}
%     \label{fig:trigram}
% \end{figure*}

% \begin{figure*}
%     \centering
%     \subfigure[\centering Munjong (\nth{5} King)]{\includegraphics[width=0.3\linewidth]{final_figures/5th_trigram.pdf}}
%     \subfigure[\centering Seonjo (\nth{14} King)]{\includegraphics[width=0.3\linewidth]{final_figures/14th_trigram.pdf}}
%     \subfigure[\centering Sunjo (\nth{23} King)]{\includegraphics[width=0.3\linewidth]{final_figures/23rd_trigram.pdf}}
%     \caption{Number of Overlapped Trigrams per King era. It shows the changes of trigrams over kings whose x-axis shows the kings and the y-axis shows the number of overlapped trigrams.}
%     \label{fig:trigram}
% \end{figure*}

\begin{figure}[!ht]
    \centering
    \subfigure[\centering Taejo (\nth{1} King)]{\includegraphics[width=0.45\columnwidth]{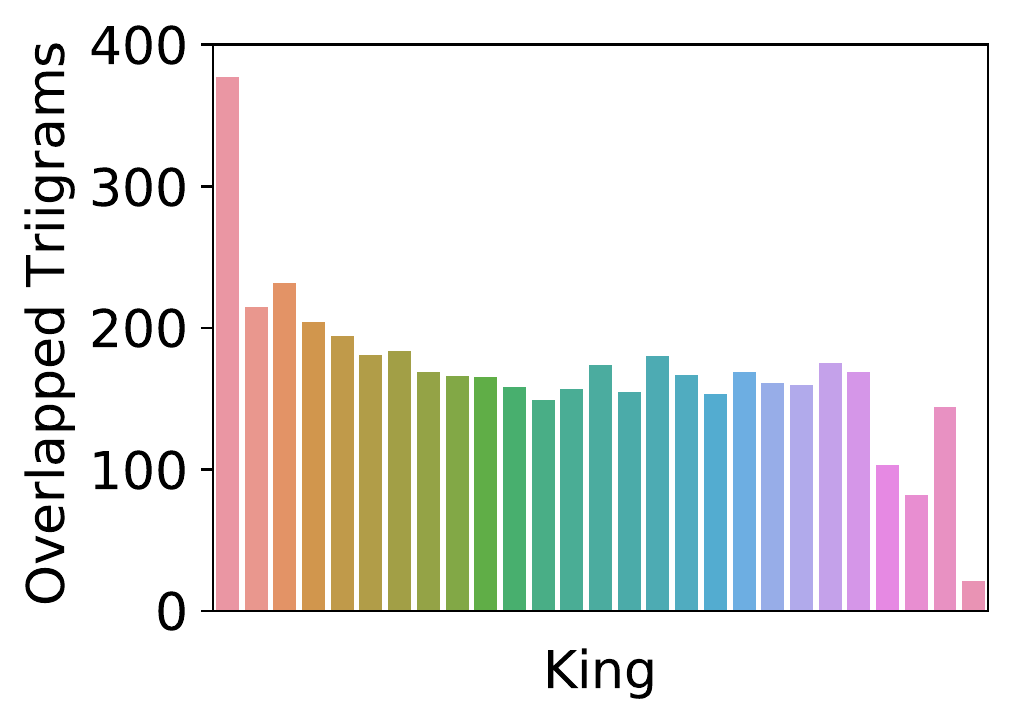}}
    \subfigure[\centering Seongjong (\nth{9} King)]{\includegraphics[width=0.45\columnwidth]{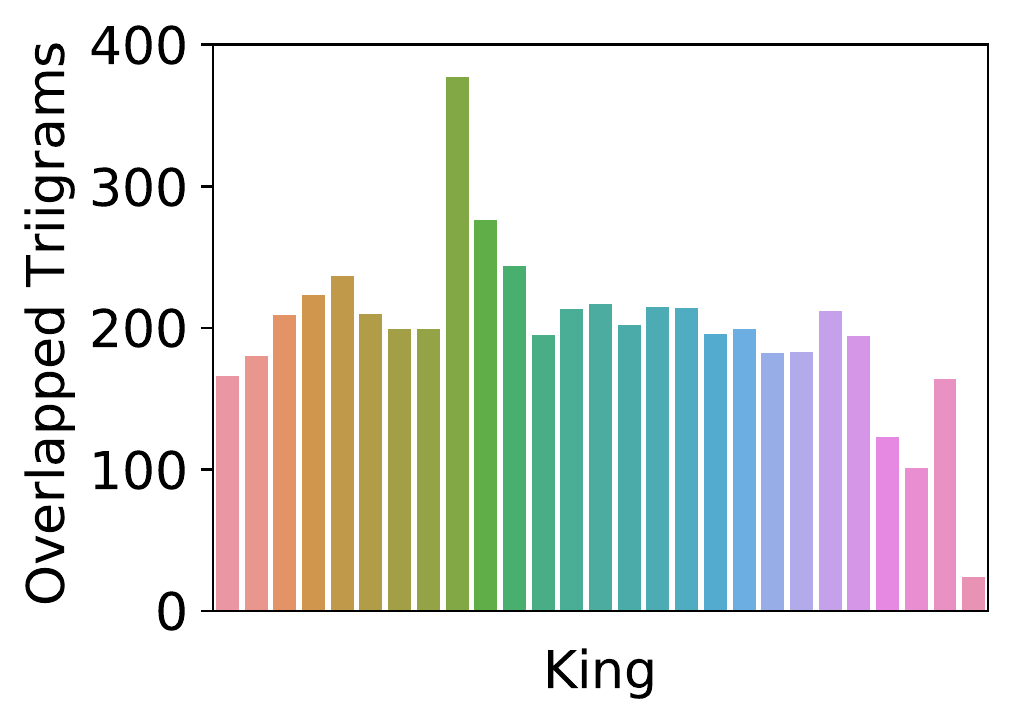}}
    \subfigure[\centering Hyojong (\nth{17} King)]{\includegraphics[width=0.45\columnwidth]{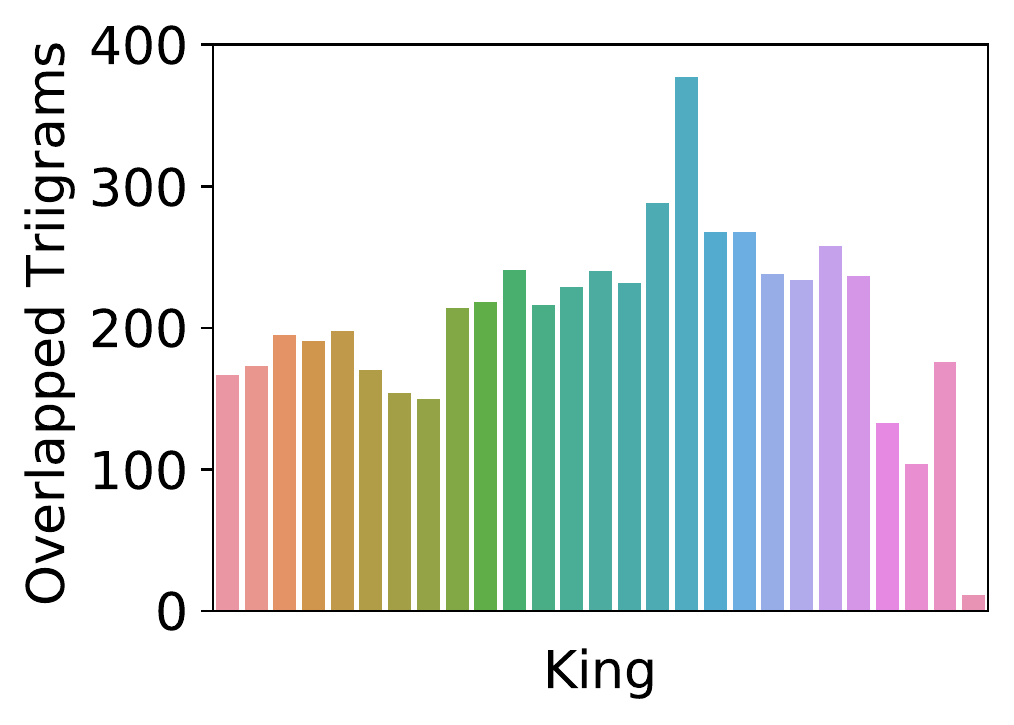}}
    \subfigure[\centering Cheoljong (\nth{25} King)]{\includegraphics[width=0.45\columnwidth]{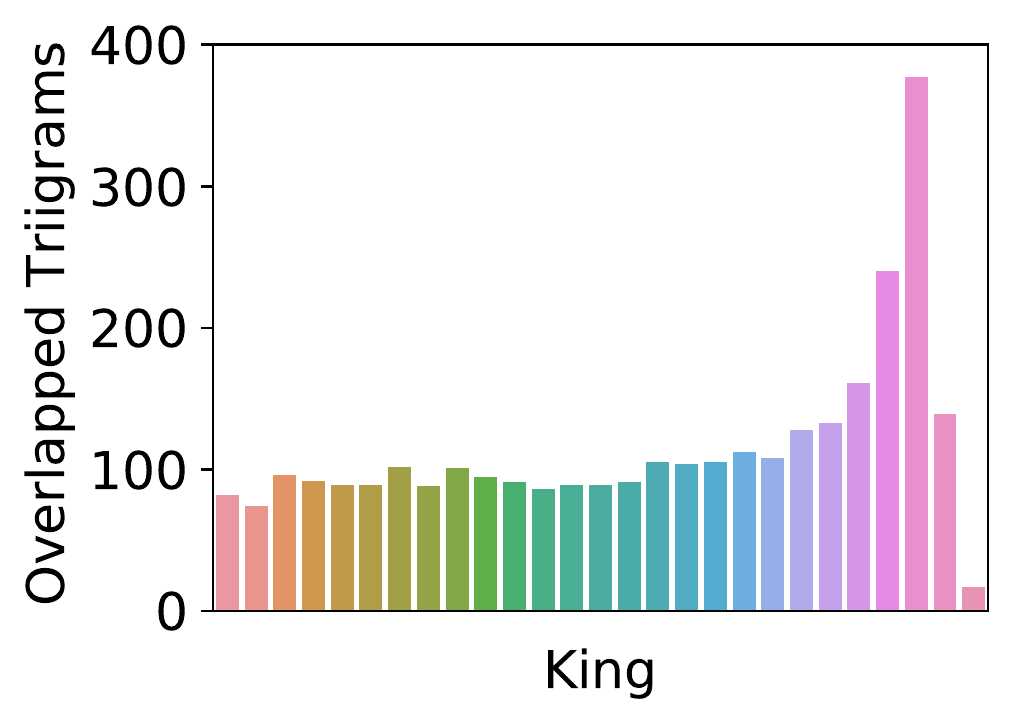}}
    \caption{Number of overlapped trigrams per king era. It shows the changes of trigrams over kings whose x-axis shows the kings and the y-axis shows the number of overlapped trigrams.}
    \label{fig:trigram}
\end{figure}

% \begin{figure*}
%     \centering
%     \subfigure[\centering Taejo (\nth{1} King)]{\includegraphics[width=0.22\linewidth]{final_figures/1st_trigram.pdf}}
%     \subfigure[\centering Seongjong (\nth{9} King)]{\includegraphics[width=0.22\linewidth]{final_figures/9th_trigram.pdf}}
%     \subfigure[\centering Hyojong (\nth{17} King)]{\includegraphics[width=0.22\linewidth]{final_figures/17th_trigram.pdf}}
%     \subfigure[\centering Cheoljong (\nth{25} King)]{\includegraphics[width=0.22\linewidth]{final_figures/25th_trigram.pdf}}
%     \caption{Number of Overlapped Trigrams per King era. It shows the changes of trigrams over kings whose x-axis shows the kings and the y-axis shows the number of overlapped trigrams.}
%     \label{fig:trigram}
% \end{figure*}

We analyze how frequently words change over time.
For each king, we plot how many trigrams overlap by each king era in the order.
Figure \ref{fig:trigram} shows overlapped trigrams in the 1st, 9th, 17th, and 25th king and the detailed results with all kings are described in Appendix.
It is consistently observed that the closer the king era is, the more trigrams are overlapped.
These changes result from not named entity but lexical choices, considering that person and location entities account for 6.38\% and 2.05\% in the characters of AJD, respectively.
It verifies that words used in Joseon dynasty had changed over time gradually, and it enables the language models to capture those features.
% Figure \ref{fig:trigram} shows the changes of trigrams over kings whose x-axis shows the kings and the y-axis shows the number of overlapped trigrams.
% It implies that words used in the Joseon dynasty had changed over time, and our models captured those features to predict the written dates.
% The same figure for all kings are described at Appendix.

\paragraph{Named Entity Changes over Time}
% \begin{figure}
%     \centering
%     \subfigure[\centering Person]{\includegraphics[width=\columnwidth]{final_figures/person.pdf}}
%     \subfigure[\centering Location]{\includegraphics[width=\columnwidth]{final_figures/location.pdf}}
%     \caption{Frequencies of top-10 named entities per king}
%     \label{fig:top-10_entity}
% \end{figure}

% \begin{figure}
%     \centering
%     \includegraphics[width=\columnwidth]{final_figures/person.pdf}
%     \caption{Frequencies of Top-10 Named Person Entities per King}
%     \label{fig:top-10_entity}
% \end{figure}

\begin{figure}[!t]
    \centering
    \subfigure[\centering Person]{\includegraphics[height=0.6\columnwidth]{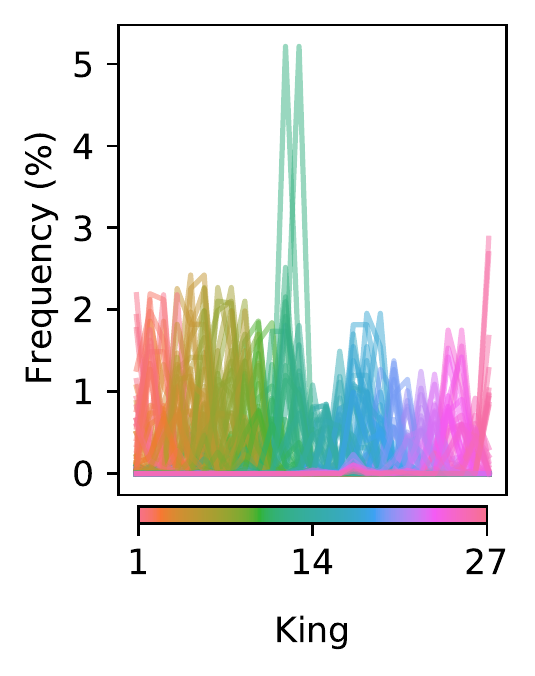}}
    \subfigure[\centering Location]{\includegraphics[height=0.6\columnwidth]{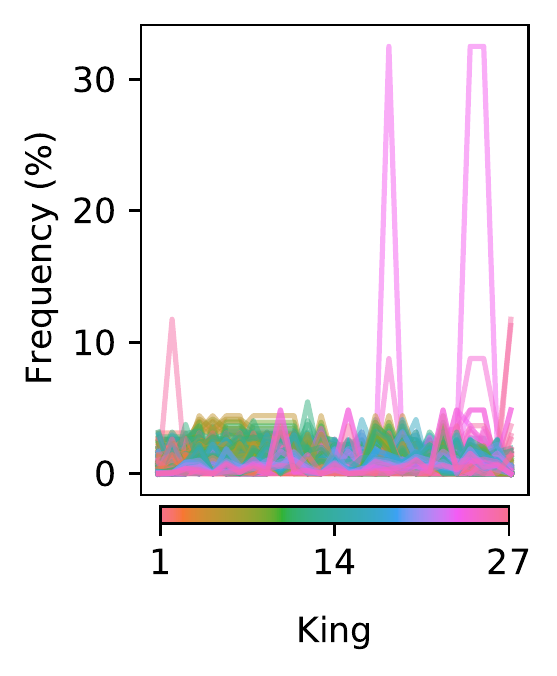}}
    \caption{Relative frequency change of top-10 named entities per king. Each line indicates the change of the relative frequency of one entity over time, and the color of the line indicates the king era in which the entity is contained in the top-10 entities. The x-axis represents the time (the kings in the order) and the y-axis represents the relative frequency in each king era.}
    \label{fig:top-10_entity}
\end{figure}
We investigate how the named entities had been used over time.
In particular, we show frequency rates of top-10 frequently-used named entities by each king era and how they change over time in Figure \ref{fig:top-10_entity}.
% In particular, we show for person entities and location entities, how the frequencies of the top-10 frequently-used named entities of each king era change over time in the 
% Figure \ref{fig:top-10_entity}.
% Figure \ref{fig:top-10_entity} shows frequencies of the top-10 named entities per each king.
It implies a strong correlation between person entity and the passage of time, while there is no explicit correlation to location entity. Most person entities include officials of the time or the previous kings, relevant to the time. In contrast, most location entities include neighboring countries or place names in the Joseon,
which are less dependent on the time.
The examples of frequently appeared named entities are described in Appendix.

\section{Related Work}

ML based NLP techniques have been recently applied to anthology to discover historical documents such as authorship attribution \citep{Ouamour2012AAAT, Sayoud2017Authorship, Reisi2020Authorship, Hossain2020Authorship}, NER \citep{Won2018NER, Palladino2020GreekNER}, and manuscript age detection \citep{Adam2018KERTAS}.
\citet{assael2022restoring} proposed Ithaca to restore ancient Greek inscriptions and perform geographical attribution and chronological attribution of them.

Along with these works, several works provide language models suited for historical texts in ancient languages and evaluate those models on existing NLU tasks, which aims to support understanding those documents considering that the target languages are mostly extinct.
\citet{Bamman2020LatinBERT} propose Latin BERT for part-of-speech tagging in ancient Latin script.
\citet{Tian2021AnchiBERT} suggest AnchiBERT and evaluate their model on some NLP tasks including poem topic classification.

However, there has been no research attempting to propose language models in Hanja, which is a dead language in Korea but absolutely necessary to explore Korean history.
Most of the studies with Hanja only shed lights on translating historical Hanja documents and use AJD as their corpus \citep{Park2020Ancient, Jin2020DWE, Kang2021Restoring}.
% \citet{Kang2021Restoring}

\section{Conclusion}
We present HUE (\textbf{H}anja \textbf{U}nderstanding \textbf{E}valuation) dataset and BERT-based pretrained language models for classical Hanja documents. HUE dataset includes diverse tasks that can support analyzing historical documents written in Hanja which is an extinct language in Korea: Chronological Attribution (CA), Topic Classification (TC), Named Entity Recognition (NER), and Summary Retrieval (SR).
Our models pretrained on Hanja corpora outperform other language models and we observe their performance on zero-shot settings with DRRI which is the dataset never been introduced in NLP community.
The experimental results in king prediction imply that our models capture the historical events or facts disclosed in the texts.
We also explore several methods to support Hanja language models such as masking named entities and giving document age as input sources, based on the analyses on textual features in AJD.
% Our models pretrained on Hanja corpora outperform others even on zero-shot settings, showing the needs of language models pretrained on corpora written in the same era in the classical domains. We analyze the changes of named entities and words over time and demonstrate the correlation between written ages and other tasks showing the significance of king prediction. 

Help of adequate resources in Hanja documents might could fill some caveats in our work which lacks additional experiments and analyses on the records of different genre such as poetry, novel, and humanities resulting from the low resources that we can exploit.
% Some caveats in our work that could be filled in with the help of adequate resources would be the lack of experiments and analyses on the records of different genre such as potery, novel and humanities due to lack of translated or preprocessed Hanja documents. 
However, we expect that our dataset and accompanying language models might facilitate future works on historical documents written in Hanja by providing fundamental resources to leverage unknown Hanja corpora.

\section*{Acknowledgements}
We would like to thank Yoonman Heo (Institute for the Translation of Korean Classics) providing expertise on hanja and Korean Classical Chinese.
This research was supported by the Engineering Research Center Program through the National Research Foundation of Korea (NRF) funded by the Korean Government MSIT (NRF-2018R1A5A1059921).
This work was partly supported by Institute of Information \& communications Technology Planning \& Evaluation (IITP) grant funded by the Korea government (MSIT) (No.2019-0-00421, Artificial Intelligence Graduate School Program (Sungkyunkwan University)).
Kyunghyun Cho was supported by the NYU Center for Data Science National Science Foundation (Award 1922658) and Samsung Advanced Institute of Technology (Next Generation Deep Learning: from pattern recognition to AI).


\bibliography{anthology,customclean}

\begin{thebibliography}{20}
\expandafter\ifx\csname natexlab\endcsname\relax\def\natexlab#1{#1}\fi

\bibitem[{Adam et~al.(2018)Adam, Baig, Al-Maadeed, Bouridane, and
  El-Menshawy}]{Adam2018KERTAS}
Kalthoum Adam, Asim Baig, Somaya Al-Maadeed, Ahmed Bouridane, and Sherine
  El-Menshawy. 2018.
\newblock \href {https://doi.org/http://dx.doi.org/10.1007/s10032-018-0312-3}
  {Kertas: dataset for automatic dating of ancient arabic manuscripts}.
\newblock \emph{International journal on Document Analysis and Recognition
  (IJDAR)}.

\bibitem[{Assael et~al.(2022)Assael, Sommerschield, Shillingford, Bordbar,
  Pavlopoulos, Chatzipanagiotou, Androutsopoulos, Prag, and
  de~Freitas}]{assael2022restoring}
Yannis Assael, Thea Sommerschield, Brendan Shillingford, Mahyar Bordbar, John
  Pavlopoulos, Marita Chatzipanagiotou, Ion Androutsopoulos, Jonathan Prag, and
  Nando de~Freitas. 2022.
\newblock Restoring and attributing ancient texts using deep neural networks.
\newblock \emph{Nature}, 603(7900):280--283.

\bibitem[{Bak and Oh(2015)}]{Bak2015Mining}
JinYeong Bak and Alice Oh. 2015.
\newblock \href {https://doi.org/10.18653/v1/W15-3702} {Five centuries of
  monarchy in {K}orea: Mining the text of the annals of the {J}oseon dynasty}.
\newblock In \emph{Proceedings of the 9th {SIGHUM} Workshop on Language
  Technology for Cultural Heritage, Social Sciences, and Humanities
  ({L}a{T}e{CH})}.

\bibitem[{Bamman and Burns(2020)}]{Bamman2020LatinBERT}
David Bamman and Patrick~J. Burns. 2020.
\newblock \href {http://arxiv.org/abs/2009.10053} {Latin bert: A contextual
  language model for classical philology}.

\bibitem[{Devlin et~al.(2019)Devlin, Chang, Lee, and
  Toutanova}]{Devlin2019BERT}
Jacob Devlin, Ming-Wei Chang, Kenton Lee, and Kristina Toutanova. 2019.
\newblock \href {https://doi.org/10.18653/v1/N19-1423} {{BERT}: Pre-training of
  deep bidirectional transformers for language understanding}.
\newblock In \emph{Proceedings of the NAACL}.

\bibitem[{Godbole and Sarawagi(2004)}]{Godbole2004hammingscore}
Shantanu Godbole and Sunita Sarawagi. 2004.
\newblock \href {https://doi.org/https://doi.org/10.1007/978-3-540-24775-3_5}
  {Discriminative methods for multi-labeled classification}.
\newblock In \emph{Advances in Knowledge Discovery and Data Mining}.

\bibitem[{Hossain et~al.(2020)Hossain, Akter, and
  Islam}]{Hossain2020Authorship}
Anika~Samiha Hossain, Nazia Akter, and Md.~Saiful Islam. 2020.
\newblock \href {https://doi.org/10.1145/3377049.3377079} {A stylometric
  approach for author attribution system using neural network and machine
  learning classifiers}.
\newblock In \emph{Proceedings of the International Conference on Computing
  Advancements}.

\bibitem[{Jin et~al.(2020)Jin, Wi, Kang, and Kim}]{Jin2020DWE}
KyoHoon Jin, JeongA Wi, KyeongPil Kang, and YoungBin Kim. 2020.
\newblock \href {https://doi.org/10.3390/app10217939} {Korean historical
  documents analysis with improved dynamic word embedding}.
\newblock \emph{Applied Sciences}.

\bibitem[{Kang et~al.(2021)Kang, Jin, Yang, Jang, Choo, and
  Kim}]{Kang2021Restoring}
Kyeongpil Kang, Kyohoon Jin, Soyoung Yang, Soojin Jang, Jaegul Choo, and
  Youngbin Kim. 2021.
\newblock \href {https://doi.org/10.18653/v1/2021.naacl-main.317} {Restoring
  and mining the records of the {J}oseon dynasty via neural language modeling
  and machine translation}.
\newblock In \emph{Proceedings of the NAACL: Human Language Technologies}.

\bibitem[{Nogueira and Cho(2019)}]{Nogueira2019Reranker}
Rodrigo Nogueira and Kyunghyun Cho. 2019.
\newblock \href {http://arxiv.org/abs/1901.04085} {Passage re-ranking with
  {BERT}}.
\newblock \emph{CoRR}.

\bibitem[{Ouamour and Sayoud(2012)}]{Ouamour2012AAAT}
Siham Ouamour and Halim Sayoud. 2012.
\newblock \href {https://doi.org/10.1109/ICCITechnol.2012.6285841} {authorship
  attribution of ancient texts written by ten arabic travelers using a smo-svm
  classifier}.
\newblock In \emph{International Conference on Communications and Information
  Technology (ICCIT)}.

\bibitem[{Palladino et~al.(2020)Palladino, Karimi, and
  Mathiak}]{Palladino2020GreekNER}
Chiara Palladino, Farimah Karimi, and Brigitte Mathiak. 2020.
\newblock \href {https://doi.org/http://dx.doi.org/10.17613/j7jt-b052} {Ner on
  ancient greek with minimal annotation}.
\newblock In \emph{https://dh2020. adho. org/}.

\bibitem[{Park et~al.(2020)Park, Lee, Yang, and Lim}]{Park2020Ancient}
Chanjun Park, Chanhee Lee, Yeongwook Yang, and Heuiseok Lim. 2020.
\newblock \href {https://doi.org/10.1109/ACCESS.2020.3004879} {Ancient korean
  neural machine translation}.
\newblock \emph{IEEE Access}.

\bibitem[{Reisi and Mahboob~Farimani(2020)}]{Reisi2020Authorship}
Ehsan Reisi and Hassan Mahboob~Farimani. 2020.
\newblock \href
  {https://doi.org/https://dx.doi.org/10.22034/jaisis.2021.269735.1018}
  {Authorship attribution in historical and literary texts by a deep learning
  classifier}.
\newblock \emph{journal of Applied Intelligent Systems and Information
  Sciences}.

\bibitem[{Robertson and Zaragoza(2009)}]{Robertson2009BM25}
Stephen Robertson and Hugo Zaragoza. 2009.
\newblock \href {https://doi.org/10.1561/1500000019} {The probabilistic
  relevance framework: Bm25 and beyond}.
\newblock \emph{The Probabilistic Relevance Framework (PRF)}.

\bibitem[{Sayoud and Ouamour(2017)}]{Sayoud2017Authorship}
Halim Sayoud and Siham Ouamour. 2017.
\newblock \href
  {https://aaai.org/ocs/index.php/FLAIRS/FLAIRS17/paper/view/15408} {Score
  fusion based authorship attribution of ancient arabic texts}.
\newblock In \emph{Florida Artificial Intelligence Research Society
  Conference}.

\bibitem[{Tian et~al.(2021)Tian, Yang, Liu, and Lv}]{Tian2021AnchiBERT}
Huishuang Tian, Kexin Yang, Dayiheng Liu, and Jiancheng Lv. 2021.
\newblock \href {https://doi.org/10.1109/IJCNN52387.2021.9534342} {Anchibert: A
  pre-trained model for ancient chinese language understanding and generation}.
\newblock In \emph{2021 International Joint Conference on Neural Networks
  (IJCNN)}.

\bibitem[{Vale~de Gato(2015)}]{Gato2015anthology}
Margarida Vale~de Gato. 2015.
\newblock \href {https://doi.org/10.1080/1750399X.2015.1011901} {The
  collaborative anthology in the literary translation course}.
\newblock \emph{The Interpreter and Translator Trainer}.

\bibitem[{Won et~al.(2018)Won, Murrieta-Flores, and Martins}]{Won2018NER}
Miguel Won, Patricia Murrieta-Flores, and Bruno Martins. 2018.
\newblock \href {https://doi.org/10.3389/fdigh.2018.00002} {Ensemble named
  entity recognition (ner): Evaluating ner tools in the identification of place
  names in historical corpora}.
\newblock \emph{Frontiers in Digital Humanities}.

\bibitem[{Youden(1950)}]{Youden1950index}
William~J Youden. 1950.
\newblock \href
  {https://doi.org/https://doi.org/10.1002/1097-0142(1950)3:1%3C32::aid-cncr2820030106%3E3.0.co;2-3}
  {Index for rating diagnostic tests}.
\newblock \emph{Cancer}, 3(1):32--35.

\end{thebibliography}
\bibliographystyle{acl_natbib}

\clearpage
\section*{Appendix}
\appendix
\section{Model}

Table \ref{tab:model_config} shows hyperparameter settings of our models.
We used Intel(R) Xeon(R) Silver 4114 (40 CPUs) and GeForce RTX 2080 Ti 10GB (4 GPUs) for all experiments including training, fine-tuning, and inference.

\begin{table}[!h]
\centering
\begin{tabular}{@{}ll@{}}
\toprule
Hyperparameter          & Value  \\ \midrule
Batch Size              & 32     \\
Early Stopping Patience & 3      \\
Hidden Size             & 768    \\
Learning Rate           & 2e-5   \\
Learning Rate Scheduler & Linear \\
Max Sequence Length     & 512    \\
Number of Hidden Layers & 12     \\
Optimizer               & AdamW  \\
Vocab Size              & 11270  \\ \bottomrule
\end{tabular}
\caption{Model configuration}
\label{tab:model_config}
\end{table}

\section{HUE Dataset}

\subsection{Dataset Size}

\begin{table}[h]
\centering
\begin{tabular}{@{}c|ccc@{}}
\toprule
    & Train   & Dev    & Test   \\ \midrule
CA  & 330,469 & 41,309 & 41,309 \\
TC  & 330,424 & 41,303 & 41,304 \\
NER & 385,915 & 13,417 & 13,418 \\
SR  & 169,840 & 21,570 & 21,296 \\ \bottomrule
\end{tabular}
\caption{Data split in HUE dataset}
\label{tab:data_size}
\end{table}

% \subsection{Class Map for Topic Classification}

\subsection{Source Corpora}

% \subsection{Motivation for Data Creation}

% \subsection{Dataset Composition}

\subsubsection{Data Collection Process}

We crawl AJD \footnote{\url{http://sillok.history.go.kr/main/main.do}}, DRS \footnote{\url{http://sjw.history.go.kr/main.do}}, and DRRI \footnote{\url{http://kyudb.snu.ac.kr/series/main.do?item_cd=ILS}} from the comprehensive database for Korean classics which are publicly available published by IKTC.
All source corpora are fully tagged with the written ages and named entities, while their entity maps differ to each other.
AJD also provides topics which is tagged by the experts in the translation process.

\subsubsection{Dataset Preprocessing}

\begin{table*}[!]
\begin{CJK*}{UTF8}{bsmi}
\begin{tabularx}{\linewidth}{@{}l|X|X@{}}
\toprule
          & Good Examples                                                                                                                                                                                                        & Bad Examples                                                                                                                                                                                                                                         \\ \midrule
Date      & King Jeongjo 17 (1793) Feb 06                                                                                                                                                                                        & King Yeongjo 50 (1774) May 15                                                                                                                                                                                                                        \\ \midrule
Gang      & 遞承旨 徐榮輔以沈晉賢代之前望也                                                                                                                                                                                                     & 行召對于尊賢閣。 兼弼善 洪景顏。說書 李啇駿。翊贊 李應重 。                                                                                                                                                                                                                     \\
Mok       & 榮輔不爲仕進政院請牌招敎以許遞前望單子入之待下批牌招察任                                                                                                                                                                                         & 講續綱目                                                                                                                                                                                                                                                 \\ \midrule
Gang (En) & Royal secretary Yeong-bo Seo was replaced and Jin-hyeon Shim was appointed.                                                                                                                                          & Crown Prince held a royal lecture at the Office of Crown Prince. Kyung-ahn Hong, a lecturer for the Crown Prince, Sangjoon Lee, the second tutor of the Office of Lectures for the Crown Prince, and Eungjoon Lee, a Guard of Crown Prince attended. \\
Mok (En)  & When Yeong-bo Seo did not resign, the king summoned his servants through his royal secretary and ordered him to do so, saying, “wait for appointing the royal secretary among candidates and let him check his job”. & They delivered the lecture on 《Sokgangmok (Comprehensive Mirror for Aid in Government)》.                                                                                                                                                             \\ \bottomrule
\end{tabularx}
\end{CJK*}
\caption{Good and bad examples in DRRI}
\label{tab:DRRI_example}
\end{table*}
% 강: 승지 서영보(徐榮輔)를 체차하고 심진현(沈晉賢)을 임명하였다
% 목: ○ 서영보가 사진(仕進)하지 않자 정원이 패초하기를 청한 데 대해 전교하기를, “체차해 주고 전망 단자를 들이며, 하비(下批)하기를 기다려 패초하여 직임을 살피게 하라.” 하였다.

% 강: 존현각에서 소대를 행하였다. 겸필선 홍경안, 설서 이상준, 익찬 이응중이 들어왔다.
% 목: ○ 《속자치통감강목》을 강하였다.

Table \ref{tab:DRRI_example} shows good and bad example in DRRI to use as summary retrieval dataset. Bad examples mostly written from the \nth{21} King Yeongjo to early in the \nth{22} King Jungjo describe daily lifes of the crown prince who is King Jeongjo. The bad example in Table \ref{tab:DRRI_example} is depicting his study. These examples tend to present extremely short \textit{mok}s which cannot be treated as summary and content, while the offical records on administrative has much longer \textit{mok}s.

% \subsection{Dataset Distribution}

% \subsection{Dataset Maintenance}

% \subsection{Legal \& Ethical Considerations}

% \subsection{Model Details}

% \subsection{Indended Use}

% \subsection{Factors}

% \subsection{Metrics}

% \subsection{Training Data}

% \subsection{Evaluation Data}

% \subsection{Quantitative Analyses}

% \subsection{Ethical Considerations}

% \subsection{Caveats and Recommendations}

\section{Discussion}

\subsection{Trigram Changes Over Time}

\begin{figure*}
    \centering
    \includegraphics[width=\linewidth]{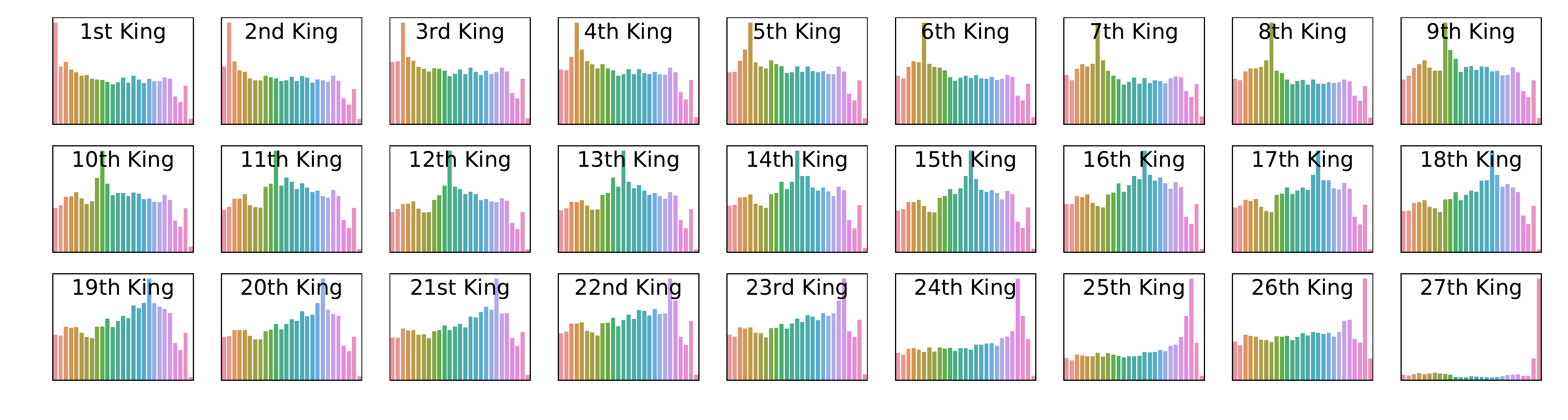}
    \caption{Trigram changes over time}
    \label{fig:bigram_all}
\end{figure*}

Figure \ref{fig:bigram_all} shows the changes of trigrams over all kings.
It clearly delivers the changes of trigrams as time goes by. We can observe the same trend in either unigram or bigram.

\subsection{Top-5 Named Entities by Kings}

\begin{table*}[!]
\resizebox{\linewidth}{!}{
\begin{CJK*}{UTF8}{bsmi}
\begin{tabular}{@{}l|lr|lr|lr@{}}
\toprule
\multicolumn{1}{c|}{\begin{tabular}[c]{@{}c@{}}King\\ (Reigning period)\end{tabular}} & \multicolumn{2}{c|}{\begin{tabular}[c]{@{}c@{}}Munjong (\nth{5} King)\\ (1450-1452)\end{tabular}} & \multicolumn{2}{c|}{\begin{tabular}[c]{@{}c@{}}Seonjo (\nth{14} King)\\ (1567-1608)\end{tabular}} & \multicolumn{2}{c}{\begin{tabular}[c]{@{}c@{}}Sunjo (\nth{23} King)\\ (1790-1834)\end{tabular}} \\ \midrule
\multirow{5}{*}{Person}                                                               & 世宗                                             & 2.42\%                                           & 柳成龍                                            & 1.08\%                                           & 南公轍                                            & 1.21\%                                         \\
                                                                                      & 金宗瑞                                            & 2.26\%                                           & 李德馨                                            & 0.81\%                                           & 金載瓚                                            & 1.09\%                                         \\
                                                                                      & 李季甸                                            & 1.97\%                                           & 尹斗壽                                            & 0.72\%                                           & 李時秀                                            & 0.85\%                                         \\
                                                                                      & 皇甫仁                                            & 1.82\%                                           & 李恒福                                            & 0.65\%                                           & 沈象奎                                            & 0.83\%                                         \\
                                                                                      & 鄭苯                                             & 1.78\%                                           & 李元翼                                            & 0.57\%                                           & 李相璜                                            & 0.77\%                                         \\
\multirow{5}{*}{Person (En)}                                                          & King Sejong                                    & (1392-1397)                                      & Seongryong Ryu                                 & (1542-1607)                                      & Gongcheol Nam                                  & (1760-1840)                                    \\
                                                                                      & Jongseo Kim                                    & (1383-1453)                                      & Deokhyeong Lee                                 & (1561-1613)                                      & Jaechan Kim                                    & (1746-1827)                                    \\
                                                                                      & Kyejeon Lee                                    & (1404-1459)                                      & Dushu Yun                                      & (1533-1601)                                      & Sisu Lee                                       & (1745-1821)                                    \\
                                                                                      & Boin Hwang                                     & (1387-1453)                                      & Hangbok Lee                                    & (1556-1618)                                      & Sangkyu Shim                                   & (1766-1838)                                    \\
                                                                                      & Bun Jeong                                      & (1394-1454)                                      & Weonik Lee                                     & (1547-1634)                                      & Sanghwang Lee                                  & (1763-1841)                                    \\ \midrule
\multirow{5}{*}{Location}                                                             & 平安                                             & 4.40\%                                           & 平壤                                             & 2.10\%                                           & 春塘臺                                            & 2.02\%                                         \\
                                                                                      & 咸吉                                             & 3.24\%                                           & 朝鮮                                             & 1.93\%                                           & 漢城府                                            & 1.19\%                                         \\
                                                                                      & 黃海                                             & 2.28\%                                           & 京畿                                             & 1.17\%                                           & 仁政殿                                            & 1.71\%                                         \\
                                                                                      & 輝德殿                                            & 2.21\%                                           & 全羅                                             & 1.76\%                                           & 平安                                             & 1.51\%                                         \\
                                                                                      & 京畿                                             & 2.10\%                                           & 慶尙                                             & 1.64\%                                           & 景慕宮                                            & 1.49\%                                         \\
\multirow{5}{*}{Location (En)}                                                        & \textbf{Pyong-an}                              & \textbf{(Province)}                              & {\ul Pyongyang}                                & {\ul (Province)}                                 & Chundangdae                                    & (Site)                                         \\
                                                                                      & Hamgyong                                       & (Province)                                       & Joseon                                         & (Country)                                        & Hanseong Magistracy                            & (Province)                                     \\
                                                                                      & {\ul Hwanghae}                                 & {\ul (Province)}                                 & \textbf{Gyeonggi}                              & \textbf{(Province)}                              & Injeongjeon                                    & (Hall)                                         \\
                                                                                      & Hwideokjeon                                    & (Hall)                                           & {\ul Jeolla}                                   & {\ul (Province)}                                 & \textbf{Pyong-an}                              & \textbf{(Province)}                            \\
                                                                                      & \textbf{Gyeonggi}                              & \textbf{(Province)}                              & {\ul Gyeongsang}                               & {\ul (Province)}                                 & Gyeongmogung                                   & (Palace)                                       \\ \bottomrule
\end{tabular}
\end{CJK*}
}
\caption{Top-5 named entities in \nth{5}, \nth{14}, and \nth{23} kings}
\label{tab:top5-entity}
\end{table*}

Table \ref{tab:top5-entity} shows top-5 person and location named entities in three King reigns. All person entities except King Sejong, the most frequent entity in Munjong, are officials in the reign periods, which is different by time changes. In contary, all location entities are the name of place, palace or site, showing some entities overlap among kings.

% \clearpage
\section{Experimental Results}

% \subsection{King Prediction}

\begin{table*}[!]
\centering
\begin{tabular}{@{}l|ccccc@{}}
\toprule
                                    & Acc               & F1                & Pre               & Rec               & QWK               \\ \midrule
BERT not pretrained                 & 56.59             & 54.26             & 55.08             & 56.59             & 87.09             \\
AnchiBERT \cite{Tian2021AnchiBERT}  & 76.31             & 75.74             & 76.45             & 76.31             & 93.99             \\
mBERT \cite{Devlin2019BERT}         & 76.02             & 75.29             & 75.80             & 76.02             & 93.76             \\
AnchiBERT + AJD/DRS                 & \textbf{79.50}    & \textbf{79.33}    & \textbf{80.06}    & \textbf{79.50}    & \textbf{95.46}    \\
mBERT + AJD/DRS                     & \underline{77.99} & \underline{77.78} & \underline{78.92} & \underline{77.99} & \underline{95.04} \\ \bottomrule
\end{tabular}
\caption{Evaluation results for our PLMs on chronological attribution}
\label{tab:king}
\end{table*}

For CA, we measure the Quadratic Weighted Kappa score (QWK score) as metrics that treat each king label hierarchically.

% \subsection{Topic Classification}

\begin{table*}[!]
\resizebox{\linewidth}{!}{
% \usepackage{booktabs}
\begin{tabular}{@{}l|ccccc|ccccc@{}}
\toprule
                                    & \multicolumn{5}{c|}{\textbf{Major (4 classes)}}                                                   & \multicolumn{5}{c}{\textbf{Minor (106 classes)}}                                                  \\
                                    & Acc               & F1                & Pre               & Rec               & Ham               & Acc               & F1                & Pre               & Rec               & Ham               \\ \midrule
BERT not pretrained                 & 68.52             & 68.91             & 74.82             & 70.65             & 79.04             & 26.48             & 61.29             & 54.86            & 84.59             & 42.45             \\
AnchiBERT \cite{Tian2021AnchiBERT}  & 68.99             & 85.81             & 85.24             & 88.44             & 83.00             & 31.47             & 69.30             & 61.63            & 90.26             & 51.63             \\
mBERT \cite{Devlin2019BERT}         & 56.78             & 79.59             & 77.60             & 87.78             & 73.64             & \underline{32.85} & 70.01             & 62.17            & 90.84             & \underline{54.28} \\
AnchiBERT + AJD/DRS                 & \textbf{70.48}    & \textbf{88.33}    & \textbf{86.61}    & \textbf{92.71}    & \textbf{84.80}    & 31.77             & \textbf{72.07}    & \textbf{64.87}    & \underline{91.24} & 52.25             \\
mBERT + AJD/DRS                     & \underline{69.15} & \underline{87.13} & \underline{85.48} & \underline{91.36} & \underline{83.81} & \textbf{33.96}    & \underline{71.28} & \underline{63.50} & \textbf{91.56}    & \textbf{55.91}    \\ \bottomrule
\end{tabular}
}
\caption{Evaluation results for our PLMs on topic classification}
\label{tab:topic}
\end{table*}

Since TC is a multi-label classification task whose example might have multiple labels as the answer, we measure Hamming score along with accuracy. In this case, accuracy is the exact match score, and the hamming score is the accuracy of subset matched, $\mid T \cap P \mid / \mid T \cup P \mid$, where $T$ is set of true labels and $P$ is set of predicted labels \cite{Godbole2004hammingscore}. For the evaluation results in Table \ref{tab:topic}, we find and set the best threshold to each label by Youden's index. All pretrained models outperform BERT without pretraining, and two LMs re-trained on hanja documents show the best performances.

% \subsection{Named Entity Recognition}

\begin{table*}[!]
\resizebox{\linewidth}{!}{
\begin{tabular}{@{}l|cccc|ccc|ccc@{}}
\toprule
                    & \multicolumn{4}{c|}{\textbf{Overall}}                                         & \multicolumn{3}{c|}{\textbf{Person}}                      & \multicolumn{3}{c}{\textbf{Location}}                     \\
                    & Acc               & F1                & Pre               & Rec               & F1                & Pre               & Rec               & F1                & Pre               & Rec               \\ \midrule
BERT not pretrained & 98.67             & 89.40             & 90.58             & 88.25             & 92.13             & 93.98             & 90.36             & 87.10             & 87.76             & 86.46             \\
AnchiBERT           & \underline{98.72} & 90.30             & \underline{90.98} & 89.62             & \underline{93.13} & \underline{94.47} & 91.83             & \underline{87.91} & \underline{88.08} & \underline{87.75} \\
mBERT               & 98.52             & 88.57             & 89.52             & 87.64             & 91.63             & 93.60             & 89.74             & 86.02             & 86.18             & 85.85             \\
AnchiBERT + AJD/DRS & \textbf{98.76}    & \textbf{90.42}    & \textbf{91.31}    & \underline{89.55} & \textbf{93.28}    & \textbf{94.77}    & \underline{91.84} & \textbf{88.01}    & \textbf{88.43}    & 87.59             \\
mBERT + AJD/DRS     & 98.69             & \underline{90.16} & 90.36             & \textbf{89.95}    & 92.83             & 93.51             & \textbf{92.17}    & 87.90             & 87.74             & \textbf{88.06}    \\ \bottomrule
\end{tabular}
}
\caption{Evaluation results for our PLMs on named entity recognition}
\label{tab:ner}
\end{table*}

% \subsection{Summary Retrieval}

\begin{table*}[!]
\centering
\begin{tabular}{@{}l|ccc@{}}
\toprule
                                    & MRR               & Top-1             & Top-10            \\ \midrule
BM25                                & 29.87             & 25.58             & 33.98             \\
BERT not pretrained                 & 52.85             & 99.20             & 99.64             \\
AnchiBERT \cite{Tian2021AnchiBERT}  & 67.92             & 99.20             & \underline{99.85} \\
mBERT \cite{Devlin2019BERT}         & 67.06             & 99.32             & 99.50             \\
AnchiBERT + AJD/DRS                 & \textbf{74.29}    & \textbf{99.64}    & \textbf{99.91}    \\
mBERT + AJD/DRS                     & \underline{73.88} & \underline{99.44} & 99.59             \\ \bottomrule
\end{tabular}
\caption{Evaluation results for our PLMs on summary retrieval}
\label{tab:summarization}
\end{table*}

\end{document}